%% file: main.tex
\documentclass{article}

\PassOptionsToPackage{numbers, compress}{natbib}

\input{preamble}

\newcommand{\name}{UniTTA}
\newcommand{\lorename}{COFA}

\usepackage[preprint]{neurips_2024}

\usepackage[utf8]{inputenc} 
\usepackage[T1]{fontenc}    
\usepackage{hyperref}       
\usepackage{url}            
\usepackage{booktabs}       
\usepackage{amsfonts}       
\usepackage{nicefrac}       
\usepackage{microtype}      
\usepackage{xcolor}         

\makeatletter
\renewcommand{\thanks}[1]{\footnotemark[1]\protected@xdef\@thanks{\@thanks%
    \protect\footnotetext[1]{#1}}}
\makeatother

\title{\name: Unified Benchmark and Versatile Framework Towards Realistic Test-Time Adaptation}

\author{%
    Chaoqun Du\ \ \ Yulin Wang\ \ \ Jiayi Guo\ \ \ Yizeng Han\ \ \ Jie Zhou\ \ \ Gao Huang\thanks{Corresponding author.} \\
    Tsinghua University \\
}

\begin{document}

\maketitle

\input{sec/0_abstract}
\input{sec/1_intro}

\input{sec/2_related}

\input{sec/3_method}

\input{sec/4_exp}

\input{sec/5_conclusion}

\bibliography{main}
\bibliographystyle{plain}


\newpage
\appendix

\input{sec/6_appendix}

\end{document}

%% file: preamble.tex
\usepackage{microtype}
\usepackage{graphicx}
\usepackage{subfigure}
\usepackage{booktabs} 

\usepackage{hyperref}

\usepackage{amsmath}
\usepackage{amssymb}
\usepackage{mathtools}
\usepackage{amsthm}

\usepackage[capitalize,noabbrev]{cleveref}

\theoremstyle{plain}

\newtheorem{proposition}{Proposition}

\newtheorem{corollary}{Corollary}
\theoremstyle{definition}
\newtheorem{definition}{Definition}

\theoremstyle{remark}

\usepackage[textsize=tiny]{todonotes}

\usepackage{bm}
\usepackage{booktabs}
\usepackage{listings}

\usepackage{extra_styles}
\usepackage{multirow}
\usepackage{array}
\usepackage{paralist}
\usepackage{comment}        %
\usepackage{bbm}
\usepackage[normalem]{ulem}
\usepackage[accsupp]{axessibility}  %
\usepackage{algorithm}
\usepackage{algorithmic}
\usepackage{wrapfig}

\crefname{section}{Sec.}{Secs.}
\Crefname{section}{Section}{Sections}
\Crefname{table}{Table}{Tables}
\crefname{table}{Tab.}{Tabs.}
\Crefname{figure}{Figure}{Figures}
\crefname{figure}{Fig.}{Figs.}
\Crefname{equation}{Equation}{Equations}
\crefname{equation}{Eq.}{Eqs.}
\Crefname{algorithm}{Algorithm}{Algorithms}
\crefname{algorithm}{Alg.}{Algs.}
\Crefname{proposition}{Proposition}{Propositions}
\crefname{proposition}{Prop.}{Props.}
\Crefname{appendix}{Appendix}{Appendices}
\crefname{appendix}{App.}{Apps.}

\usepackage{xcolor}
\usepackage{colortbl}
\definecolor{mygray}{gray}{.9}

{ \bgroup
  \addtolength\abovedisplayshortskip{#1}
  \addtolength\abovedisplayskip{#1}
  \addtolength\belowdisplayshortskip{#1}
  \addtolength\belowdisplayskip{#1}}
{\egroup\ignorespacesafterend}

%% file: sec/0_abstract.tex
\begin{abstract}

Test-Time Adaptation (TTA) aims to adapt pre-trained models to the target domain during testing.
In reality, this adaptability can be influenced by multiple factors.
Researchers have identified various challenging scenarios and developed diverse methods to address these challenges, such as dealing with continual domain shifts, mixed domains, and temporally correlated or imbalanced class distributions.
Despite these efforts, a unified and comprehensive benchmark has yet to be established.
To this end, we propose a \textbf{Uni}fied \textbf{T}est-\textbf{T}ime \textbf{A}daptation (\textbf{\name}) benchmark, which is comprehensive and widely applicable.
Each scenario within the benchmark is fully described by a Markov state transition matrix for sampling from the original dataset.
The \name~benchmark considers both domain and class as two independent dimensions of data and addresses various combinations of imbalance/balance and i.i.d./non-i.i.d./continual conditions, covering a total of \( (2 \times 3)^2 = 36 \) scenarios.
It establishes a comprehensive evaluation benchmark for realistic TTA and provides a guideline for practitioners to select the most suitable TTA method.
Alongside this benchmark, we propose a versatile \name~framework, which includes a Balanced Domain Normalization (BDN) layer and a COrrelated Feature Adaptation (\lorename) method--designed to mitigate distribution gaps in domain and class, respectively.
Extensive experiments demonstrate that our \name~framework excels within the \name~benchmark and achieves state-of-the-art performance on average.
Our code is available at \url{https://github.com/LeapLabTHU/UniTTA}.

\end{abstract}

%% file: sec/1_intro.tex
\section{Introduction}
\label{sec:intro}

Deep learning has achieved significant success across various tasks~\cite{Krizhevsky2012,LeCun2015,He2016,densenet,huang2019densenet,swin,convnext}.
However, the performance of models often degrade due to domain shifts of test data in practical deployments~\cite{ganin2015unsupervised,cycada,kundu2020universal,li2020model,liang2020we,QuinoneroCandela2022,Du2024,Du2024a}.
To mitigate this issue, a new line of research called Test-Time Adaptation (TTA)~\cite{Iwasawa2021,Bateson2022,Gandelsman2022,Zhou2023,Guo2024} has emerged, focusing on adapting models to the target domain of test data.
The core method of TTA involves using shifted domain normalization statistics to recalibrate the model's Batch Normalization (BN) layers~\cite{Ioffe2015}, addressing the mismatch between the normalization statistics of BN layer and target domain.
Early TTA methods assumed that test data is independent and identically distributed (i.i.d.), primarily re-estimating normalization statistics based on the current test batch~\cite{Nado2020, Schneider2020, Wang2021}.
In reality, scenarios are often more complex, and the i.i.d. assumption may not hold~\cite{Boudiaf2022}. 
Moreover, domain shifts can be more severe~\cite{Wang2022}, necessitating robust adaptation methods.

Recent studies have extended TTA to more realistic scenarios, proposing various methods to address challenges such as continual domain shifts~\cite{Wang2022}, mixed domains~\cite{Marsden2024,Tomar2024}, and temporally correlated~\cite{Boudiaf2022,Gong2022,Yuan2023} or imbalanced class distributions~\cite{Su2024}.
As representative examples, NOTE~\cite{Gong2022} introduces an instance-aware BN method to adjust normalization for temporally correlated data.
Balanced BN is proposed in TRIBE~\cite{Su2024} to achieve unbiased estimation of statistics, aiming to address imbalanced class distributions.
UnMIX-TNS~\cite{Tomar2024} recalibrates the statistics of each sample by multiple distinct statistics components.
However, many of the current methods have only been evaluated in specific scenarios and lack a unified and comprehensive benchmark for performance assessment.

To address this issue, we propose a \textbf{Uni}fied \textbf{T}est-\textbf{T}ime \textbf{A}daptation (\textbf{\name}) benchmark that is both comprehensive and widely applicable.
We present a novel method for constructing test data of various scenarios using a defined Markov state transition matrix.
The \name~benchmark can assist researchers in evaluating their methods in a more comprehensive and realistic manner, facilitating the development of versatile and robust TTA methods.
Moreover, it also provides a evaluating benchmark for practitioners to select the most suitable TTA method for their specific scenarios.

Based on the \name~benchmark, we conduct a comprehensive evaluation of existing methods, which reveals that these methods are not universally effective across realistic scenarios.
To obtain a versatile and robust TTA method, we need to simultaneously address domain and class distribution shifts.
This poses two primary challenges for BN recalibration: potential domain non-i.i.d. and imbalance leading to inaccurate domain-wise statistics, and class non-i.i.d. and imbalance further biasing domain-wise statistics towards majority classes.

In this work, we simultaneously tackle both challenges by proposing a novel Balanced Domain Normalization (BDN) layer.
Our primary insight is to unify both domain-aware and class-aware normalization.
We compute the statistics for each class within each domain and then average across classes to obtain balanced domain-wise statistics, mitigating the impact of class imbalance on domain-wise statistics.
During prediction, we select the corresponding statistics based on the current sample's domain, effectively addressing domain non-i.i.d. and imbalance.
Unlike class-aware normalization, which can directly utilize pseudo-labels from network output, domain information is agnostic, lacking effective criteria for determining a sample's domain or even the number of domains.
Our method is inspired by the observation that the Kullback-Leibler (KL) divergence between normalization statistics $\mu$ and $\sigma$ can effectively indicate domain-wise distribution shifts.
We utilize the KL divergence between the instance statistics and the domain-wise statistics to determine the domain of the sample.
Based on this, we also propose a dynamic method to adaptively expand the domains, which regards the KL divergence between the instance statistics and the BN layer statistics of pre-trained model as the expansion criterion.

Moreover, to address potential temporal correlation of class, we leverage the correlation characteristic by referencing the feature of the previous sample, resulting in an effective and efficient method named \lorename~(COrrelated Feature Adaptation), without requiring any modifications to model parameters.
However, a direct implementation of \lorename~may lead to performance degradation in i.i.d. scenarios.
To solve this issue, we propose a confidence-based filtering method to determine the appropriate application of \lorename.
By filtering out samples with low confidence, we ensure that \lorename~is applied only when necessary, thus improving the overall performance.

In summary, our contributions are as follows: 1) We propose a \name~benchmark, a unified and comprehensive evaluation benchmark for Realistic TTA.
2) We introduce a versatile \name~framework consisting of a Balanced Domain Normalization (BDN) layer and a \lorename~method, designed to address domain and class distribution shifts, respectively. They are simple and effective without additional training.
3) Extensive experiments based on the \name~benchmark demonstrate that our method excels in various realistic scenarios and achieves state-of-the-art performance on average.

%% file: sec/2_related.tex
\section{Related Work}
\label{sec:related}

\textbf{Test-Time Adaptation (TTA)} addresses distributional shifts in test data without requiring additional data acquisition or labeling. Sun et al.\cite{Sun2020} propose an on-the-fly adaptation method using an auxiliary self-supervised task. Subsequent TTA algorithms~\cite{Nado2020, Schneider2020, Wang2021} leverage batches of test samples to recalibrate Batch Normalization (BN) layers\cite{Ioffe2015} using test data. These studies show that using test batch statistics in BN layers can enhance robustness against distributional shifts. TENT~\cite{Wang2021} refines this approach by adapting a pre-trained model to test data through entropy minimization~\cite{Grandvalet2004}, updating a few trainable parameters in BN layers.

\textbf{Realistic Test-Time Adaptation.} Recent studies on Test-Time Adaptation (TTA) have investigated more realistic scenarios, addressing distribution changes in test data. These studies consider factors such as domain distribution shift~\cite{Wang2022,Brahma2023}, temporal correlation~\cite{Boudiaf2022, Gong2022}, and combinations of both~\cite{Yuan2023, Marsden2024, Su2024, Tomar2024}. A comprehensive comparison of these realistic settings is provided in \cref{sec:existing}. 
The methods employed in these studies include self-training~\cite{Wang2022,Yuan2023,Brahma2023}, which integrates semi-supervised self-training techniques~\cite{Huang2022} to enhance model performance, parameter-free methods~\cite{Boudiaf2022} utilizing Laplacian regularization, and Batch Normalization (BN) recalibration~\cite{Gong2022, Mirza2022, Zou2022, Yuan2023, Tomar2024, Sun2020}.
RoTTA~\cite{Yuan2023} introduces robust BN, estimating global statistics via exponential moving average.
TRIBE~\cite{Su2024} proposes a balanced BN (BBN) layer, consisting of multiple category-wise BN layers for unbiased statistic estimation. UnMIX-TNS~\cite{Tomar2024} unmixes correlated batches into $K$ distinct components, each reflecting statistics from similar test inputs.
Among these methods, BBN and UnMIX-TNS are the most similar to our work. 
However, both BBN and UnMIX-TNS consider the influence of category and domain distributions on statistics separately, which significantly limits their applicability.
In contrast, our approach simultaneously accounts for both category and domain distributions by introducing a unified BDN layer to address their combined impact on statistics.

%% file: sec/3_method.tex
\section{Benchmark}\label{sec:benchmark}

\begin{figure}
    \centering
    \includegraphics[width=1.0\linewidth]{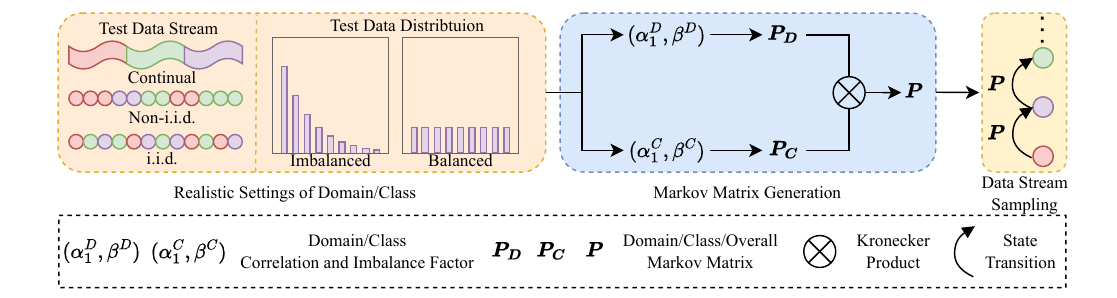}
    \vspace{-0.5cm}
    \caption{Data generation process for the \name~benchmark.
    Continual TTA describes a scenario in which the domain remains consistent over an extended period before shifting to a new domain, which exemplifies an extreme case of non-i.i.d. settings.
    We consider the domain and class as two independent attributes, each associated with its own Markov matrix.
}
    \label{fig:benchmark}
\end{figure}

In this section, we first review and analyze existing realistic TTA settings in \cref{sec:existing}.
Next, we propose a unified \name~benchmark in \cref{sec:our_benchmark}.
Finally, we discuss the advantages of our benchmark in \cref{sec:discussion_benchmark}, particularly its scalability to more complex scenarios.

\subsection{Existing Realistic TTA Settings}\label{sec:existing}

The realistic TTA settings can be divided into two categories: domain setting and class setting, as shown in \cref{tab:benchmark}.
For the class setting, real-world data streams are typically highly correlated (non-i.i.d.), which means that data categories do not change abruptly.
LAME~\cite{Boudiaf2022} introduces the concept of Non-i.i.d. TTA. to address this issue, focusing on scenarios where the domain remains constant.
For the domain setting, real-world environments can change over time, such as weather changes in autonomous driving scenarios.
CoTTA~\cite{Wang2022} introduces the concept of Continual TTA, where the domain remains stable for an extended duration before shifting to a new one.
RoTTA~\cite{Yuan2023} integrates these two settings and proposed Practical TTA, though it overlooks the data imbalance.
Subsequently, TRIBE~\cite{Su2024} proposes GLI-TTA, which further complements the Practical TTA.
Additionally, ROID~\cite{Marsden2024} further expands the domain setting by introducing the concept of Mixed Domain, where the data domain is not static but changes randomly.

Given these scenarios, we can classify the factors in existing realistic TTA settings into two categories: Temporal Correlation and Imbalance.
Therefore, a more general realistic TTA setting should consider different combinations of these factors to better simulate real-world scenarios.
Based on this analysis, a natural question arises: \emph{how can we generate such a data stream?}

\begin{table*}[!t]
    \centering
    \caption{Comparison of the proposed \name~benchmark with existing realistic TTA settings.}
    \label{tab:benchmark}
    \resizebox{1.0\linewidth}{!}{
        \begin{tabular}{llcccc}
             \toprule
           &  & \multicolumn{2}{c}{Domain Setting} & \multicolumn{2}{c}{Class Setting} \\

           \cmidrule(lr){3-4} \cmidrule(lr){5-6}
            Realistic TTA Setting  & Method      & Temporal Correlation & Imbalance & Temporal Correlation & Imbalance \\
                   \cmidrule(r){1-1}  \cmidrule(r){2-2} \cmidrule(lr){3-4} \cmidrule(lr){5-6}
            Non-i.i.d. TTA~\cite{Boudiaf2022} &LAME &  N/A (Single) & N/A (Single) & Non-i.i.d. & Imbalanced/Balanced \\
            Continual TTA~\cite{Wang2022} &CoTTA & Continual & Balanced & i.i.d. & Balanced \\
            Practical TTA~\cite{Yuan2023}& RoTTA & Continual & Balanced & Non-i.i.d. & Balanced \\
            GLI-TTA~\cite{Su2024} &TRIBE & Continual & Balanced & Non-i.i.d. & Imbalanced/Balanced \\
            Mixed Domain~\cite{Marsden2024} & ROID & i.i.d. & Balanced & i.i.d & Balanced \\
        \midrule
        \cellcolor{lightgray!50}\name~Benchmark & \cellcolor{lightgray!50}\name & \cellcolor{lightgray!50}Continual/Non-i.i.d./i.i.d. & \cellcolor{lightgray!50}Imbalanced/Balanced & \cellcolor{lightgray!50}Continual/Non-i.i.d./i.i.d. & \cellcolor{lightgray!50}Imbalanced/Balanced \\
        \bottomrule
        \end{tabular}
    }
    \vspace{-0.5cm}
\end{table*}

\subsection{\name~Benchmark}\label{sec:our_benchmark}

As previously mentioned, to efficiently construct unified datasets that adhere  to various realistic TTA settings, we propose a new \name~benchmark, based on a Markov state transition matrix from a novel local perspective.
In the following discussion, we consider the temporal correlation and imbalance of domains and classes as two independent factors.
Specifically, \emph{the Markov state can represent either the domain or the class of the data}.

Our key idea is to generate data that satisfies temporal correlation by controlling the probability of samples transitioning to themselves.
While this method might appear to neglect the issue of data imbalance, we have discovered that by properly configuring the Markov state transition matrix, \emph{we can effectively address both temporal correlation and imbalance simultaneously}.

First, we define a simple uniformly leaving Markov state transition matrix $\bm{P}$, where each element $P_{ij}$ represents the probability of transitioning from state $i$ to state $j$.
Intuitively, this transition matrix implies that the probability of transitioning from any state to any other state is uniform.

\begin{definition}[Uniformly Leaving Markov Matrix]
    A \textbf{Uniformly Leaving Markov Matrix (ULMM)} is a transition matrix in a Markov chain where each non-diagonal entry $P_{ij}$, representing the transition probability from state $i$ to state $j$ (where $i \neq j$), is identical across all states $j$. Specifically, the matrix is defined as:
    \begin{equation}
        \bm{P} =
        \begin{pmatrix}
            P_{11} & \frac{1-P_{11}}{n-1} & \cdots & \frac{1-P_{11}}{n-1} \\
            \frac{1-P_{22}}{n-1} & P_{22} & \cdots & \frac{1-P_{22}}{n-1} \\
            \vdots & \vdots & \ddots & \vdots \\
            \frac{1-P_{nn}}{n-1} & \frac{1-P_{nn}}{n-1} & \cdots & P_{nn}
        \end{pmatrix}.
    \end{equation}
\end{definition}

Based on the above definition, the ULMM can be characterized by a single vector \( \bm{\alpha} \), where \( \alpha_i = P_{ii} \).

Depending on the value of \( \alpha_i \), we can distinguish the following cases:
\begin{enumerate}
    \item When \( \alpha_i = \frac{1}{n} \), the transition probability is uniform. Consequently, the data sampled from this matrix for the \( i \)-th state is i.i.d..
    \item When \( \alpha_i = 1 \), the transition probability is inescapable, meaning that the data will remain in the \( i \)-th state until the transition probability changes, thus exhibiting continual TTA.
    \item When \( \alpha_i > \frac{1}{n} \), the probability of transitioning to itself is greater than that to other states. Thus, the data in the \( i \)-th state will exhibit temporal correlation (non-i.i.d.).
    \item When \( \alpha_i < \frac{1}{n} \), the data in the \( i \)-th state will exhibit temporal anti-correlation characteristics. However, this scenario is rare in practice and is therefore not considered in our benchmark.
\end{enumerate}

In summary, this matrix has $n$ degrees of freedom. By adjusting \(\alpha_i \in \left[\frac{1}{n}, 1\right]\), we can generate data with varying levels of temporal correlation.
Therefore, we refer to \(\bm{\alpha}\) as the (temporal) correlation vector and \(\alpha_i\) as the {(temporal) correlation factor}.

A key question we address is {whether the state distribution of data sampled from a ULMM satisfies the criteria for imbalance}.
According to Markov Chain theory, {this distribution corresponds to the stationary distribution of the matrix}, as stated in the following proposition:

\begin{proposition}[Stationary Distribution]
    For a Uniformly Leaving Markov Matrix with diagonal elements $\bm{\alpha}$ where $\alpha_i = P_{ii}$ for all $i$, there exists a unique stationary distribution $\bm{\pi} = (\pi_1, \pi_2, \cdots, \pi_n)$. This distribution satisfies the following relationship:
    \begin{equation}
       (1-\alpha_1)\pi_1 = (1-\alpha_2)\pi_2 = \cdots = (1-\alpha_n)\pi_n.
        \label{eq:stationary}
    \end{equation}
    \label{prop:stationary}
\end{proposition}
\vspace{-0.5cm}

To ensure that the sampled data follows a {long-tail distribution} (assuming, without loss of generality, that\(\frac{\pi_1}{\pi_2} = \cdots = \frac{\pi_{n-1}}{\pi_n} \geq 1\), where \(\frac{\pi_1}{\pi_n} = \beta\) is the {imbalance factor}), the configurations of \(\bm{\alpha}\) are described by the following corollary:

\begin{corollary}[Temporal Correlation and Imbalance]
    If the category distribution of data sampled based on a Uniformly Mixing Markov Matrix follows a long-tailed (power law) distribution characterized by an imbalance factor $\beta \geq 1$. Under these conditions, $\bm{\alpha}$ are constrained such that:
    \begin{equation}
        \frac{1-\alpha_1}{1-\alpha_n} = \frac{1}{\beta}, \quad \text{and}  \quad \frac{1-\alpha_1}{1-\alpha_2} = \frac{1-\alpha_2}{1-\alpha_3} = \cdots = \frac{1-\alpha_{n-1}}{1-\alpha_n} = \left(\frac{1}{\beta}\right)^{\frac{1}{n-1}}.
        \label{eq:imbalance}
    \end{equation}
    Additionally, if the distribution exhibits temporal correlation, which implies that $\alpha_1, \alpha_2, \ldots, \alpha_n > \frac{1}{n}$, then the following inequality holds between $\alpha_1$ and $\beta$:
    \begin{equation}
        (1-\alpha_1)\beta < \frac{n-1}{n}.
        \label{eq:temporal}
    \end{equation}
    \label{cor:imbalance}
\end{corollary}
\vspace{-0.5cm}

In summary, as shown in \cref{fig:benchmark}, generating data that satisfies both temporal correlation and imbalance requires tuning two parameters of the ULMM.
\emph{Specifically, it is sufficient to set the (maximum temporal) correlation factor $\alpha_1\in[1/n,1]$ and the imbalance factor $\beta\in[1,\infty)$ to satisfy \cref{eq:temporal}}.
The remaining $\alpha_i$ can then be determined by \cref{eq:imbalance}.
Thus, four parameters are needed to define two ULMMs for domain and class.
We then combine the domain and class ULMMs using the Kronecker product to obtain a final ULMM for sampling, where the (domain, class) pair is treated as a new state.

Moreover, to generate data where $\alpha_1$ and $\beta$ do not satisfy \cref{eq:temporal}, such as in the case of generating i.i.d. and imbalanced data with $\alpha_1=1/n$ and $\beta > 1$, We can pre-calculate the number of samples required for each state based on $\beta$.
During the sampling process, if the number of samples for a particular state reaches the predetermined limit, the corresponding column of the ULMM is modified to prevent further sampling of that state.

\paragraph{Sampling Time.} A practical concern is the time required for the sampling process.
Given that the ULMM matrix remains stationary most of the time (except for certain special data points, as previously discussed), we can pre-sample a sequence of transitions for each state before the actual sampling begins.
During the sampling process, we can then directly use these pre-sampled sequences, significantly reducing the time needed.
This approach ensures that the ULMM-based data generation method does not result in significantly higher time costs compared to existing methods.

\subsection{Discussion}\label{sec:discussion_benchmark}

In this section, we discuss the scalability of the \name~benchmark.
By independently generating domain and class ULMMs, we can create a comprehensive ULMM for sampling.
Moreover, the sampling ULMM can be enhanced by considering the relationships between domains and classes.
This allows us to construct domain-dependent class ULMMs, where the transition probability of a class depends on the current domain, and vice versa.
Additionally, the ULMM can be adapted for various scenarios, such as temporal anti-correlation scenarios, non-uniform scenarios where transition probabilities to other states are unequal, and higher-order Markov Chains, where transition probabilities depend on multiple previous states, not just the current one.
In summary, the data generation method defined by the \name~benchmark is \emph{highly flexible and can be efficiently extended to meet the requirements of real-world scenarios}.

\section{\name~Framework}\label{sec:method}

In this section, we first introduce the overall framework in \cref{sec:framework}.
We then discuss its two core components: Balanced Domain Normalization layer (\cref{sec:bdn}) and the \lorename~method (\cref{sec:lore}).

\subsection{Overview}\label{sec:framework}

\begin{figure}
    \centering
    \includegraphics[width=0.8\linewidth]{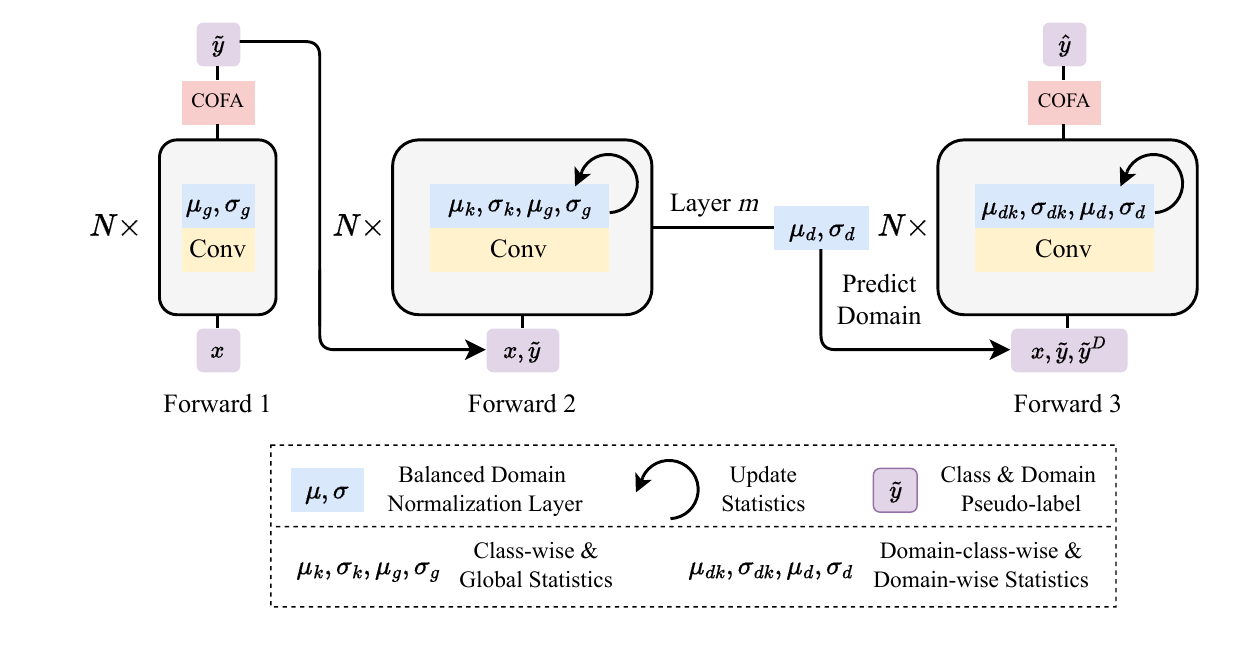}
    \vspace{-0.3cm}
    \caption{The overall architecture of the \name~framework. The original model's BN layers are replaced by BDN layers, and the linear classifier is equipped with the \lorename~method. The \name~framework sequentially predicts the class label and domain label in the $m$-th BDN layer through three forward passes, ultimately providing the final prediction.}
    \label{fig:framework}
    \vspace{-0.3cm}
\end{figure}

As illustrated in the \cref{fig:framework}, our framework substitutes the original BN with our BDN layer and equips the linear classifier with our \lorename~method.
The \name~framework utilizes a progressive prediction strategy through three forward passes:
\begin{itemize}
    \item \textbf{Forward 1:} In the absence of prior domain and class information, we perform a forward pass using global statistics to obtain initial pseudo-labels.
    \item \textbf{Forward 2:} With class labels available, we conduct a second forward pass, updating both class and global statistics. At a specified BDN layer (as a hyper-parameter), we also predict the domain based on domain statistics.
    \item \textbf{Forward 3:} Finally, with both class and domain labels, we perform a forward pass using domain statistics for the final prediction, updating both domain-class and domain statistics.
\end{itemize}

\subsection{Balanced Domain Normalization}\label{sec:bdn}

The core idea of Balanced Domain Normalization (BDN) is to implement a domain-aware normalization in an unsupervised manner.
To counteract the bias caused by imbalanced class data, which skews domain statistics towards the majority classes, we suggest calculating both domain-specific and class-specific statistics.
By averaging these statistics, we can remove the class bias and obtain more accurate domain statistics.

For each sample, we calculate the instance statistics~\cite{Ulyanov2016}, which are essential for domain assignment, expansion, and updating the statistics.
The instance statistics are defined as follows:
\begin{equation}
    \small
\bm{\mu}_i = \frac{1}{HW} \sum_{h=1}^{H} \sum_{w=1}^{W} F_{c,h,w}, \quad \bm{\sigma}^2_i = \frac{1}{HW} \sum_{h=1}^{H} \sum_{w=1}^{W} (F_{c,h,w} - \bm{\mu}_i)^2.
\end{equation}
Here, \( F_{c,h,w} \) represents the feature map of the \( c \)-th channel at position \((h,w)\), and \( H \) and \( W \) denote the height and width of the feature map, respectively.

\paragraph{Domain assignment (prediction) and expansion.}

First, all domain statistics, including those generated by expansions, are initialized using the corresponding batch normalization (BN) statistics of the original pretrained model ($\bm{\mu}_\text{ori}$, $\bm{\sigma}^2_\text{ori}$). Initially, the number of domains is set to one.

Next, {domain assignment and the decision to expand the domain are performed at a specific layer, which is the only hyper-parameter in our method}.
Specifically, we calculate the Kullback-Leibler (KL) divergence between the instance statistics of each sample and the domain statistics, assuming they follow a normal distribution. 
If the  KL divergence of the sample to all domain statistics is \emph{greater than that to the original domain statistics, the sample is considered to belong to a new domain}, necessitating domain expansion during the Forward 3.
This condition is satisfied when:
\begin{equation}
    \min_{d} D^S_{\text{KL}}(\mathcal{N}(\bm{\mu}_i, \bm{\sigma}^2_i) \,||\, \mathcal{N}(\bm{\mu}_d, \bm{\sigma}^2_d)) > D^S_{\text{KL}}(\mathcal{N}(\bm{\mu}_i, \bm{\sigma}^2_i) \,||\, \mathcal{N}(\bm{\mu}_\text{ori}, \bm{\sigma}^2_\text{ori})),
\end{equation}
where $D^S_{\text{KL}}$ is the symmetric KL divergence, defined as:
\begin{equation}
    \small
    D^S_{\text{KL}}(\mathcal{N}(\bm{\mu}_i, \bm{\sigma}^2_i) \,||\, \mathcal{N}(\bm{\mu}_d, \bm{\sigma}^2_d)) = D_{\text{KL}}(\mathcal{N}(\bm{\mu}_d, \bm{\sigma}^2_d) \,||\, \mathcal{N}(\bm{\mu}_i, \bm{\sigma}^2_i)) + D_{\text{KL}}(\mathcal{N}(\bm{\mu}_i, \bm{\sigma}^2_i) \,||\, \mathcal{N}(\bm{\mu}_d, \bm{\sigma}^2_d)).
\end{equation}
Otherwise, the sample is assigned to the domain with the minimum KL divergence:
\begin{equation}
    \hat{y}^D_i = \argmin_{d} D^S_{\text{KL}}(\mathcal{N}(\bm{\mu}_i, \bm{\sigma}^2_i) \,||\, \mathcal{N}(\bm{\mu}_d, \bm{\sigma}^2_d)).
\end{equation}

\paragraph{Domain-class statistics update.}

Based on the domain assignment, we update the domain-class statistics ($\bm{\mu}_{dk}$, $\bm{\sigma}_{dk}$) and domain statistics ($\bm{\mu}_d$, $\bm{\sigma}_d$) using the instance statistics $\bm{\mu}_i$ and $\bm{\sigma}^2_i$.
The class statistics ($\bm{\mu}_k$, $\bm{\sigma}_k$) and global statistics ($\bm{\mu}_g$, $\bm{\sigma}_g$) can be considered as the domain-class statistics and domain statistics for a single domain, respectively.
Various updating methods can be applied independently of our core method.
We utilize the commonly applied Exponential Moving Average (EMA) to update the domain-class statistics $\bm{\mu}_{dk}$ and $\bm{\sigma}_{dk}$.
Specifically, we adopt the EMA update from Balanced BN~\cite{Su2024} without modification.
For detailed update rules, please refer to~\cref{sec:appendix_bdn}.

\subsection{\lorename}\label{sec:lore}

The \lorename~method leverages the correlation characteristics of data to enhance prediction accuracy by utilizing the information of the previous sample when predicting the current sample.
Implementing this method is straightforward, requiring only the storage of feature from the previous sample.
Specifically, the classifier with \lorename~is defined as follows:
\begin{equation}
\bm{p}^\text{\lorename}_i = \text{softmax} \left( \frac{\bm{w}^T (\bm{z}_i + \bm{z}_{i-1})}{2} + b \right),
\end{equation}
where $\bm{{z}}_i$ is the feature of the $i$-th sample, and $\bm{w}$ and $b$ are the weight and bias of the  original classifier of  the pre-trained model, respectively.
However, direct implementation of \lorename~results in a marked performance decrease under i.i.d. conditions
To address this, we propose a confidence filtering strategy to combine the predictions of the \lorename~and the original classifier, as follows:
\begin{equation}
\bm{p}_i = \begin{cases}
\bm{p}^\text{\lorename}_i, & \text{if}\ \max(\bm{p}^\text{\lorename}_i) > \max(\bm{p}^\text{single}_i) \\
\bm{p}^\text{single}_i, & \text{otherwise}
\end{cases}
\end{equation}
where $\bm{p}^\text{single}_i = \text{softmax}(\bm{w}^T \bm{z}_i + b)$.
By filtering out low-confidence predictions, this approach balances performance in both i.i.d. and non-i.i.d. conditions, as shown in \cref{sec:appendix_more_results}.

Furthermore, to enhance the robustness  of the \name~framework, \emph{we also implement the confidence filtering that considers the prediction results of the Forward 2 and 3}.

%% file: sec/4_exp.tex
\begin{table*}[!t]
    \caption{\textbf{Average error ($\%$) on CIFAR10-C within the \name~benchmark.}
    $(\{i,n,1\},\{1,u\})$ denotes correlation and imbalance settings, where $\{i,n,1\}$ represent i.i.d., non-i.i.d. and continual, respectively, and $\{1,u\}$ represent balance and imbalance, respectively.
    Corresponding setting denotes the existing setting and method as shown in \cref{tab:benchmark}.
}
\label{tab:main_cifar10}
\centering
 \resizebox{1.0\linewidth}{!}{%
     \begin{tabular}{lccccccccccccl}

     \toprule

     Class setting   & \multicolumn{2}{c}{i.i.d. and balanced (i,1)} & \multicolumn{5}{c}{non-i.i.d. and balanced (n,1)}  & \multicolumn{5}{c}{non-i.i.d. and imbalanced (n,u)} \\

     \cmidrule(r){1-1} \cmidrule(lr){2-3} \cmidrule(lr){4-8} \cmidrule(lr){9-13}

     Domain setting    &  (1,1)  & (i,1)   & (1,1) & (i,1) & (i,u) & (n,1) & (n,u) & (1,1) & (i,1) & (i,u) & (n,1) & (n,u)\\

     \cmidrule(r){1-1} \cmidrule(lr){2-3} \cmidrule(lr){4-8} \cmidrule(lr){9-13}
     Corresponding setting &  CoTTA  & ROID & RoTTA & -- & -- & -- & -- & TRIBE & -- & -- & -- & -- & Avg.\\
     \cmidrule(r){1-1} \cmidrule(lr){2-3} \cmidrule(lr){4-8} \cmidrule(lr){9-13} \cmidrule(l){14-14}

     TENT~\cite{Wang2021}        &  {24.03}         &  {59.37}         &  {70.29}  &  {83.23}  &  {73.16}  &  {78.79}  &  {69.18}  &  {47.40}         &  {59.57}  &  {51.48}  &  {62.00}  &  {51.90} &  {60.87} \\
CoTTA~\cite{Wang2022}       &  \textbf{16.68}         &  {33.76}         &  {53.21}  &  {63.93}  &  {62.77}  &  {61.67}  &  {61.21}  &  {41.46}         &  {55.18}  &  {50.27}  &  {53.20}  &  {50.42} &  {50.31} \\
BN~\cite{Nado2020}          &  {21.00}         &  {34.18}         &  {49.42}  &  {57.00}  &  {54.82}  &  {56.26}  &  {54.91}  &  {41.52}         &  {50.65}  &  {46.52}  &  {50.08}  &  {47.27} &  {46.97} \\
ROID~\cite{Marsden2024}     &  {16.92}  &  {31.00}         &  {43.79}  &  {56.93}  &  {53.78}  &  {53.55}  &  {52.75}  &  {41.35}         &  {52.55}  &  {48.47}  &  {51.43}  &  {48.84} &  {45.95} \\
TEST                        &  {43.46}         &  {43.45}         &  {43.76}  &  {43.52}  &  {40.37}  &  {43.45}  &  {40.68}  &  {42.46}         &  {42.83}  &  {38.74}  &  {42.29}  &  {39.39} &  {42.03}  \\
LAME~\cite{Boudiaf2022}     &  {45.07}         &  {44.60}         &  {41.40}  &  {40.48}  &  {36.98}  &  {40.15}  &  {37.49}  &  {40.91}         &  {40.64}  &  {36.58}  &  {40.16}  &  {36.93} &  {40.12} \\
UnMIX-TNS~\cite{Tomar2024}  &  {24.53}         &  {32.82}         &  {24.68}  &  {32.99}  &  {29.03}  &  {32.72}  &  {29.15}  &  {27.60}         &  {35.81}  &  {31.88}  &  {36.44}  &  {32.48} &  {30.84} \\
NOTE~\cite{Gong2022}        &  {22.55}         &  \textbf{24.48}  &  {31.79}  &  {38.52}  &  {27.58}  &  {32.98}  &  {28.49}  &  {34.92}         &  {34.79}  &  {28.58}  &  {33.99}  &  {30.32} &  {30.75} \\
RoTTA~\cite{Yuan2023}       &  {17.84}         &  {33.45}         &  {19.52}  &  {36.89}  &  {31.49}  &  {35.66}  &  {31.58}  &  {20.39}         &  {36.24}  &  {31.67}  &  {36.33}  &  {32.46} &  {30.29} \\
TRIBE~\cite{Su2024}         &  {18.20}         &  {31.90}         &  {18.54}  &  {32.37}  &  {28.34}  &  {32.57}  &  {28.87}  &  \textbf{17.75}  &  {32.60}  &  {28.69}  &  {32.92}  &  {29.32}  &  {27.67} \\
                            \midrule
\cellcolor{lightgray!50}\name &\cellcolor{lightgray!50}{28.38} & \cellcolor{lightgray!50}{31.34} & \cellcolor{lightgray!50}\textbf{16.40} & \cellcolor{lightgray!50}\textbf{18.53} & \cellcolor{lightgray!50}\textbf{16.19} & \cellcolor{lightgray!50}\textbf{20.09} & \cellcolor{lightgray!50}\textbf{17.20} & \cellcolor{lightgray!50}{17.93} & \cellcolor{lightgray!50}\textbf{20.88} & \cellcolor{lightgray!50}\textbf{18.46} & \cellcolor{lightgray!50}\textbf{22.89} & \cellcolor{lightgray!50}\textbf{19.88} & \cellcolor{lightgray!50}\textbf{20.68 \color{teal}{(-6.99)}} \\
    \bottomrule
    \end{tabular}
}
\vskip -0.2in
\end{table*}

\begin{table*}[!t]
    \caption{\textbf{Average error ($\%$) on CIFAR100-C within the \name~benchmark.}
    $(\{i,n,1\},\{1,u\})$ denotes correlation and imbalance settings, where $\{i,n,1\}$ represent i.i.d., non-i.i.d. and continual, respectively, and $\{1,u\}$ represent balance and imbalance, respectively.
    Corresponding setting denotes the existing setting and method as shown in \cref{tab:benchmark}.
}
\label{tab:main_cifar100}
\centering
 \resizebox{1.0\linewidth}{!}{%
     \begin{tabular}{lccccccccccccl}

     \toprule

     Class setting   & \multicolumn{2}{c}{i.i.d. and balanced (i,1)} & \multicolumn{5}{c}{non-i.i.d. and balanced (n,1)}  & \multicolumn{5}{c}{non-i.i.d. and imbalanced (n,u)} \\

     \cmidrule(r){1-1} \cmidrule(lr){2-3} \cmidrule(lr){4-8} \cmidrule(lr){9-13}

     Domain setting    &  (1,1)  & (i,1)   & (1,1) & (i,1) & (i,u) & (n,1) & (n,u) & (1,1) & (i,1) & (i,u) & (n,1) & (n,u)\\

     \cmidrule(r){1-1} \cmidrule(lr){2-3} \cmidrule(lr){4-8} \cmidrule(lr){9-13}
     Corresponding setting &  CoTTA  & ROID & RoTTA & -- & -- & -- & -- & TRIBE & -- & -- & -- & -- & Avg.\\
     \cmidrule(r){1-1} \cmidrule(lr){2-3} \cmidrule(lr){4-8} \cmidrule(lr){9-13} \cmidrule(l){14-14}

  TENT~\cite{Wang2021}        &  {81.06}         &  {91.05}          &  {96.53}  &  {97.08}  &  {94.26}  &  {95.08}  &  {89.75}         &  {93.79}  &  {93.74}  &  {86.80}  &  {88.26}  &  {83.97}  & {90.95} \\
  BN~\cite{Nado2020}          &  {36.20}         &  {46.48}          &  {76.55}  &  {79.33}  &  {78.55}  &  {79.15}  &  {79.42}         &  {64.42}  &  {69.33}  &  {69.68}  &  {69.11}  &  {68.49}  & {68.06} \\
  CoTTA~\cite{Wang2022}       &  {32.74}         &  {43.02}          &  {78.26}  &  {79.95}  &  {78.91}  &  {78.89}  &  {79.38}         &  {65.68}  &  {68.56}  &  {69.58}  &  {68.07}  &  {68.46}  & {67.62} \\
  ROID~\cite{Marsden2024}     &  \textbf{29.91}  &  \textbf{36.84}   &  {71.09}  &  {77.57}  &  {76.80}  &  {76.14}  &  {76.47}         &  {55.27}  &  {63.21}  &  {63.88}  &  {62.70}  &  {63.38}  & {62.77} \\
  NOTE~\cite{Gong2022}        &  {65.96}         &  {63.07}          &  {79.69}  &  {67.67}  &  {57.69}  &  {59.75}  &  {54.37}         &  {71.52}  &  {58.86}  &  {55.52}  &  {57.55}  &  {54.25}  & {62.16} \\
  RoTTA~\cite{Yuan2023}       &  {33.46}         &  {46.54}          &  {38.95}  &  {53.80}  &  {52.30}  &  {53.25}  &  {52.55}         &  {37.79}  &  {54.99}  &  {53.89}  &  {55.36}  &  {53.63}  & {48.88} \\
  TEST                        &  {46.35}         &  {46.43}          &  {46.64}  &  {46.66}  &  {45.11}  &  {47.33}  &  {44.89}         &  {47.07}  &  {46.86}  &  {45.83}  &  {47.87}  &  {46.05}  & {46.42} \\
  UnMIX-TNS~\cite{Tomar2024}  &  {38.94}         &  {46.32}          &  {39.12}  &  {46.88}  &  {45.66}  &  {46.92}  &  {45.36}         &  {40.19}  &  {47.44}  &  {46.41}  &  {47.55}  &  {46.20}  & {44.75} \\
  TRIBE~\cite{Su2024}         &  {33.10}         &  {45.73}          &  {34.69}  &  {47.95}  &  {43.75}  &  {47.89}  &  {44.37}         &  {32.74}  &  {46.67}  &  {43.19}  &  {46.35}  &  {44.37}  & {42.57} \\
  LAME~\cite{Boudiaf2022}     &  {48.21}         &  {47.47}          &  {34.07}  &  {32.80}  &  {30.44}  &  {33.19}  &  {29.84}  &  {37.44}  &  {36.43}  &  {34.75}  &  {37.08}  &  {34.86}  & {36.38} \\
  \midrule
   \cellcolor{lightgray!50}\name & \cellcolor{lightgray!50}{44.17} & \cellcolor{lightgray!50}{48.86} & \cellcolor{lightgray!50}\textbf{24.49} & \cellcolor{lightgray!50}\textbf{28.99} & \cellcolor{lightgray!50}\textbf{28.57} & \cellcolor{lightgray!50}\textbf{31.85} & \cellcolor{lightgray!50}\textbf{29.53} & \cellcolor{lightgray!50}\textbf{25.81} & \cellcolor{lightgray!50}\textbf{30.96} & \cellcolor{lightgray!50}\textbf{30.95} & \cellcolor{lightgray!50}\textbf{32.87} & \cellcolor{lightgray!50}\textbf{32.16} & \cellcolor{lightgray!50}\textbf{32.43 \color{teal}{(-3.95)}} \\

        \bottomrule
    \end{tabular}
}
\vskip -0.2in
\end{table*}

\begin{table*}[!t]
    \caption{\textbf{Average error ($\%$) on ImageNet-C within the \name~benchmark.}
    $(\{i,n,1\},\{1,u\})$ denotes correlation and imbalance settings, where $\{i,n,1\}$ represent i.i.d., non-i.i.d. and continual, respectively, and $\{1,u\}$ represent balance and imbalance, respectively.
    Corresponding setting denotes the existing setting and method as shown in \cref{tab:benchmark}.
}

    \label{tab:main_imagenet}
\centering
 \resizebox{1.0\linewidth}{!}{%
     \begin{tabular}{lccccccccccccl}

     \toprule

     Class setting   & \multicolumn{2}{c}{i.i.d. and balanced (i,1)} & \multicolumn{5}{c}{non-i.i.d. and balanced (n,1)}  & \multicolumn{5}{c}{non-i.i.d. and imbalanced (n,u)} \\

     \cmidrule(r){1-1} \cmidrule(lr){2-3} \cmidrule(lr){4-8} \cmidrule(lr){9-13}

     Domain setting    &  (1,1)  & (i,1)   & (1,1) & (i,1) & (i,u) & (n,1) & (n,u) & (1,1) & (i,1) & (i,u) & (n,1) & (n,u)\\

     \cmidrule(r){1-1} \cmidrule(lr){2-3} \cmidrule(lr){4-8} \cmidrule(lr){9-13}
     Corresponding setting &  CoTTA  & ROID & RoTTA & -- & -- & -- & -- & TRIBE & -- & -- & -- & -- & Avg.\\
     \cmidrule(r){1-1} \cmidrule(lr){2-3} \cmidrule(lr){4-8} \cmidrule(lr){9-13} \cmidrule(l){14-14}

  TENT~\cite{Wang2021}        &  {70.58}         &  {91.88}  &  {98.72}  &  {99.31}         &  {99.53}  &  {99.12}  &  {99.32}         &  {97.50}  &  {99.22}  &  {99.13}  &  {97.03}  &  {98.86}  &  {95.85}  \\
  ROID~\cite{Marsden2024}     &  \textbf{60.67}  &  {79.18}  &  {98.51}  &  {99.71}         &  {99.84}  &  {99.52}  &  {99.61}         &  {91.76}  &  {99.77}  &  {99.57}  &  {98.15}  &  {99.37}  &  {93.80}  \\
  NOTE~\cite{Gong2022}        &  {91.62}         &  {88.18}  &  {93.67}  &  {95.27}         &  {96.82}  &  {95.00}  &  {95.81}         &  {92.49}  &  {95.93}  &  {95.41}  &  {88.93}  &  {95.05}  &  {93.68}  \\
  CoTTA~\cite{Wang2022}       &  {66.87}         &  {80.67}  &  {95.13}  &  {96.80}         &  {97.33}  &  {96.22}  &  {96.33}         &  {89.70}  &  {95.20}  &  {94.50}  &  {92.11}  &  {93.71}  &  {91.22}  \\
  TRIBE~\cite{Su2024}         &  {75.85}         &  {84.78}  &  {89.78}  &  {92.62}         &  {96.54}  &  {95.19}  &  {95.99}         &  {88.72}  &  {92.85}  &  {93.71}  &  {89.37}  &  {94.05}  &  {90.79}  \\
  BN~\cite{Nado2020}          &  {69.33}         &  {82.87}  &  {93.79}  &  {95.08}         &  {95.15}  &  {95.10}  &  {95.01}         &  {88.40}  &  {92.24}  &  {92.25}  &  {91.31}  &  {91.84}  &  {90.20}  \\
  UnMIX-TNS~\cite{Tomar2024}  &  {79.64}         &  {85.55}  &  {79.74}  &  {84.42}         &  {82.67}  &  {84.57}  &  {82.91}         &  {78.67}  &  {83.28}  &  {82.34}  &  {85.04}  &  {82.38}  &  {82.60}  \\
  TEST                        &  {81.99}         &  {82.05}  &  {81.92}  &  {82.10}         &  {81.66}  &  {81.96}  &  {81.74}         &  {81.60}  &  {81.21}  &  {81.42}  &  {81.20}  &  {81.52}  &  {81.70}  \\
  RoTTA~\cite{Yuan2023}       &  {67.77}         &  {79.91}  &  {71.72}  &  {80.54}         &  {79.65}  &  {80.30}  &  {79.63}         &  {68.74}  &  {78.26}  &  {77.94}  &  {79.78}  &  {78.36}  & {76.88}  \\
  LAME~\cite{Boudiaf2022}     &  {82.55}         &  {82.26}  &  {74.48}  &  {72.21}  &  {71.77}  &  {73.52}  &  {73.13}  &  {75.70}  &  {73.44}  &  {73.54}  &  {74.38}  &  {74.39}  &  {75.12}  \\
  \midrule

   \cellcolor{lightgray!50}\name & \cellcolor{lightgray!50}{78.07} & \cellcolor{lightgray!50}\textbf{78.00} & \cellcolor{lightgray!50}\textbf{70.25} & \cellcolor{lightgray!50}\textbf{66.83} & \cellcolor{lightgray!50}\textbf{66.42} & \cellcolor{lightgray!50}\textbf{68.29} & \cellcolor{lightgray!50}\textbf{68.05} & \cellcolor{lightgray!50}\textbf{72.02} & \cellcolor{lightgray!50}\textbf{65.68} & \cellcolor{lightgray!50}\textbf{66.87} & \cellcolor{lightgray!50}\textbf{68.48} & \cellcolor{lightgray!50}\textbf{67.58} & \cellcolor{lightgray!50}\textbf{69.71 \color{teal}{(-5.41)}} \\

   \bottomrule
   \end{tabular}
}
\vskip -0.2in
\end{table*}

\section{Experiments}
\label{sec:exp}

In this section, we first present the main results on our proposed \name~benchmark in \cref{sec:main_results}.
Additional analysis, including ablation studies, hyperparameter sensitivity analysis and visualizations, are provided in \cref{sec:analysis}.
For detailed information on the experimental setup, please refer to \cref{sec:appendix_exp_setup}.

\subsection{Results}
\label{sec:main_results}

As discussed in \cref{sec:benchmark}, our \name~benchmark can accommodate various combinations of correlation and imbalance settings, including i.i.d./non-i.i.d./continual, balanced/imbalanced configurations, which results in $6$ possible combinations.
To better simulate real-world scenarios, we exclude the continual setting for classes, as it is rare for all samples from a single class to appear consecutively in practice.
Therefore, we consider a total of $24$ realistic settings, formed by pairing the $6$ domain settings with the $4$ class settings.
We present the results for $12$ of these settings in the main paper, encompassing both existing and the most challenging scenarios.
\emph{Additional results for all methods and components of all 24 settings for all three datasets are available in \cref{sec:appendix_more_results}}.

\paragraph{Main results and comparison with existing methods.}

We can compare the robustness of different methods across various datasets and settings in \cref{tab:main_cifar10}, \cref{tab:main_cifar100}, and \cref{tab:main_imagenet}.
Each method typically performs best in the context for which it was designed, such as CoTTA~\cite{Wang2022} and ROID~\cite{Marsden2024}.
However, these methods generally do not generalize well to other scenarios.
Notably, many approaches perform worse compared to a vanilla test on CIFAR100-C and ImageNet-C, highlighting the need for a comprehensive benchmark.
Our method outperforms the others on all datasets across most settings, consistently achieving superior performance, particularly in more realistic scenarios.

\paragraph{Evaluation under more correlation/imbalance factors.}

\begin{figure}[t]
	\centering
    \includegraphics[width=1.0\linewidth]{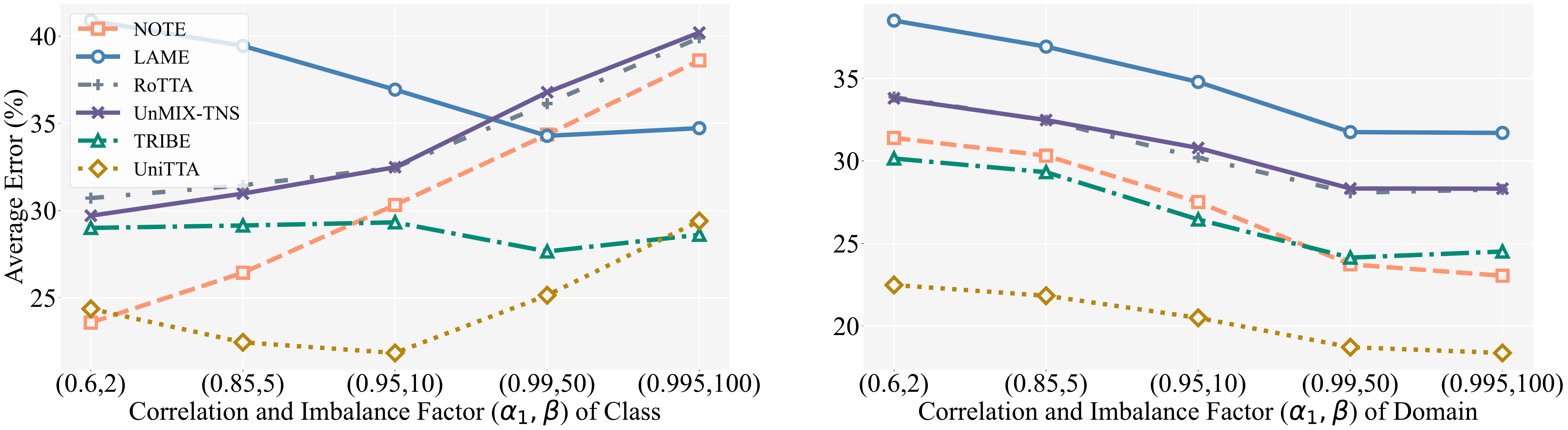}
    \vskip -0.1in
    \caption{Average error (\%) on CIFAR10-C under various correlation and imbalance factors. The default factors for domain and class are (0.85, 5) and (0.95, 10), respectively. In two sets of experiments, we kept either the domain or class factors constant while varying the other.
    }
    \label{fig:diff_factors}
\end{figure}

Additional experiments are conducted under varying correlation and imbalance factors as shown in \cref{fig:diff_factors}.
The settings are both non-i.i.d. and imbalanced in terms of domain and class distribution.
The results indicate that our method remains robust across different correlation and imbalance factors.

\subsection{Analysis}\label{sec:analysis}

\paragraph{Ablation Study.}

We conduct an ablation study across various settings and datasets to evaluate the impact of different components, benchmarking them against similar methods as shown in \cref{tab:ablation} and \cref{tab:components}.
This section presents the overall results, while detailed results are provided in \cref{sec:appendix_more_results}.

\noindent (a) \emph{Effectiveness of different components.}
We first investigate the impact of different components on model performance across all settings and datasets.
The results in \cref{tab:ablation} demonstrate the effectiveness of our two core components, \lorename~and BDN.
Additionally, applying the confidence filter further enhances model performance.

\noindent (b) \emph{Comparison with similar methods.}
We compare our two components with both parameter-free method which do not require modifications to model parameters  and normalization methods.
Our BDN consistently outperforms other normalization methods, including UnMIX-TNS~\cite{Tomar2024} and Balanced BN~\cite{Su2024}.
Notably, our \lorename~achieves performance comparable to LAME by leveraging the temporal correlation characteristic (just averaging with the last feature).
Additionally, \lorename's performance remains entirely independent of batch size, whereas LAME's performance is significantly influenced by batch size, as shown in \cref{fig:bs}.

\begin{table}[t]
  \centering
  \begin{minipage}{0.48\textwidth}
    \centering
    \caption{Ablation study of different components. The average of $12$ settings are reported on CIFAR10-C, CIFAR100-C, and ImageNet-C.}
    \resizebox{\textwidth}{!}{
    \begin{tabular}{lcccc}
    \toprule[1pt]
                                  & C10-C & C100-C & IN-C & Avg. \\
    \cmidrule(r){1-1} \cmidrule(lr){2-4} \cmidrule(l){5-5}
        TEST  & {42.03} & {46.42} & {81.70} & {56.72}\\
        \lorename (w/o filter) & {38.60} & {38.25} & {76.04} & {50.96}\\
        \lorename & {37.22} & {37.34} & {76.38} & {50.31}\\
        BN~\cite{Nado2020} &{46.97} & {68.06} & {90.20} & {68.41}\\
        BDN (w/o filter) & {27.04} & {42.79} & {76.39} & {48.74}\\
        BDN & {26.64} & {40.88} & {77.15} & {48.22}\\
    \midrule
        \cellcolor{lightgray!50}\name &\cellcolor{lightgray!50}\textbf{20.68} &\cellcolor{lightgray!50}\textbf{32.43} &\cellcolor{lightgray!50}\textbf{69.71} &\cellcolor{lightgray!50}\textbf{40.94}\\

    \bottomrule[1pt]
    \end{tabular}
    }
    \label{tab:ablation}
  \end{minipage}
  \hspace{0.5ex}
  \begin{minipage}{0.50\textwidth}
    \centering
    \caption{Comparison of our two components with parameter-free and normalization methods.}
    \resizebox{\textwidth}{!}{

    \begin{tabular}{lcccc}
    \toprule[1pt]
                                  & C10-C & C100-C & IN-C & Avg. \\
    \cmidrule(r){1-1} \cmidrule(lr){2-4} \cmidrule(l){5-5}
    \textit{Parameter-free Method} &            &            &            &      \\
    LAME~\cite{Boudiaf2022} & {40.12} & \textbf{36.38} & \textbf{75.12} & {50.74}\\
    \midrule
    \cellcolor{lightgray!50}\lorename &\cellcolor{lightgray!50}\textbf{37.22} &\cellcolor{lightgray!50}{37.34} &\cellcolor{lightgray!50}{76.38} &\cellcolor{lightgray!50}\textbf{50.31}\\
    \midrule
    \textit{Normalization Method} &            &            &            &      \\
    Robust BN~\cite{Yuan2023} & {32.34} & {46.33} & {85.30} & {54.66}\\
    UnMIX-TNS~\cite{Tomar2024} & {30.84} & {44.75} & {82.60} & {52.73}\\
    Balanced BN~\cite{Su2024} & {30.10} & {43.83} & {82.54} & {52.17}\\
    \midrule
    \cellcolor{lightgray!50}BDN &\cellcolor{lightgray!50}\textbf{26.64} &\cellcolor{lightgray!50}\textbf{40.88} &\cellcolor{lightgray!50}\textbf{77.15} &\cellcolor{lightgray!50}\textbf{48.22}\\

    \bottomrule[1pt]
    \end{tabular}

    }
    \label{tab:components}
  \end{minipage}
  \vskip -0.2in
\end{table}

\paragraph{Hyperparameter Sensitivity.}

As demonstrated in \cref{sec:appendix_hyper}, we also conduct experiments to assess the sensitivity to hyperparameters.
\cref{fig:bs} shows the performance of several competitive baselines and our method under different batch sizes.
Our method's performance remains unaffected by batch size, which can be attributed to the inherent characteristics of the BDN and \lorename~methods.
In contrast, batch-based methods such as LAME and NOTE exhibit significant sensitivity to batch size.

Our framework has only one hyperparameter: the position of the BDN for domain prediction.
The results in \cref{fig:bn} show that the performance of BDN is optimal when the first layer of an intermediate block is selected.
This also indicates that the network retains more of the original image information in the shallow layers while learning more class-specific features in the deeper layers.

\paragraph{Visualization of dynamic domain expansion.}
We also visualize the domain expansion process in \cref{fig:vis} of \cref{sec:appendix_vis}.
The process demonstrates that the BDN layer effectively captures the domain information and dynamically expands domains, which is crucial for accurate domain prediction.

%% file: sec/5_conclusion.tex
\section{Conclusion}
\label{sec:conclusion}

In this work, we propose a unified benchmark, \name, for Test-Time Adaptation (TTA), which is comprehensive and broadly applicable.
It sets a benchmark for evaluating realistic TTA scenarios and provides a guideline for selecting the most suitable TTA method for specific scenarios.
Building on this, we introduce a versatile \name~framework that simultaneously addresses domain and class distribution shifts.
Specifically, the framework includes a Balanced Domain Normalization (BDN) layer and a \lorename~method, which are simple and effective without additional training.
Empirical evidence from the \name~benchmark demonstrates that our \name~framework excels in various Realistic TTA scenarios and achieves state-of-the-art performance on average.

%% file: sec/6_appendix.tex
\section{Appendix: Method}
\label{sec:appendix_method}

This section provides the detailed proofs and the statistic update rules of BDN.

\subsection{Proof of Proposition~\ref{prop:stationary}}
\label{sec:appendix_proof}

\begin{proof}[Proof of \cref{prop:stationary}]
    By the convergence properties of Markov chains~\cite{Ross1995}, a Uniformly Leaving Markov Matrix (ULMM) \( \bm{P} \) has a unique stationary distribution \(\bm{\pi} = (\pi_1, \pi_2, \ldots, \pi_n)\) which satisfies \(\bm{\pi} = \bm{\pi}\bm{P}\).
    To solve this, we must find the nontrivial solution to the linear equation \((P^T - I)\bm{\pi} = \bm{0}\), where \(I\) is the identity matrix and \(\bm{\pi}\) is a column vector.
    Thus, we have
    \begin{equation}
        \begin{aligned}
        \begin{pmatrix}
            \alpha_1 - 1 & \frac{1-\alpha_2}{n-1} & \cdots & \frac{1-\alpha_n}{n-1} \\
            \frac{1-\alpha_1}{n-1} & \alpha_2 - 1 & \cdots & \frac{1-\alpha_n}{n-1} \\
            \vdots & \vdots & \ddots & \vdots \\
            \frac{1-\alpha_1}{n-1} & \frac{1-\alpha_2}{n-1} & \cdots & \alpha_n - 1
        \end{pmatrix}
        \begin{pmatrix}
            \pi_1 \\
            \pi_2 \\
            \vdots \\
            \pi_n
        \end{pmatrix}
        &=
        \begin{pmatrix}
            0 \\
            0 \\
            \vdots \\
            0
        \end{pmatrix}, \\
        \begin{pmatrix}
            -1 & \frac{1}{n-1} & \cdots & \frac{1}{n-1} \\
            \frac{1}{n-1} & -1 & \cdots & \frac{1}{n-1} \\
            \vdots & \vdots & \ddots & \vdots \\
            \frac{1}{n-1} & \frac{1}{n-1} & \cdots & -1
        \end{pmatrix}
        \begin{pmatrix}
            (1-\alpha_1)\pi_1 \\
            (1-\alpha_2)\pi_2 \\
            \vdots \\
            (1-\alpha_n)\pi_n
        \end{pmatrix}
        &=
        \begin{pmatrix}
            0 \\
            0 \\
            \vdots \\
            0
        \end{pmatrix}.
        \end{aligned}
    \end{equation}
    
    Observing that each row of the coefficient matrix sums to zero, there exists a non-trivial solution \(\bm{1} = (1, 1, \ldots, 1)\). Hence,
    \begin{equation}
        (1-\alpha_1)\pi_1 = (1-\alpha_2)\pi_2 = \cdots = (1-\alpha_n)\pi_n
    \end{equation}
    is one of the non-trivial solutions. By the uniqueness of the stationary distribution, the proof is complete.
\end{proof}

\subsection{Statistic Update Rules of BDN} \label{sec:appendix_bdn}

Before introducing the statistical update rules of BDN, we define a mean notation to simplify the expressions:
\begin{equation}
    \overline{F_{c,\cdot,\cdot}} = \frac{1}{HW}\sum_{h=1}^{H}\sum_{w=1}^{W}F_{c,h,w}
\end{equation}
which denotes the average over the omitted dimensions.
Using this definition, we can simplify instance statistics as follows:
\begin{equation}
\bm{\mu}_i = \overline{F_{c,\cdot,\cdot}}, \quad \bm{\sigma}^2_i = \overline{(F_{c,\cdot,\cdot} - \bm{\mu}_i)^2}.
\end{equation}
We adopt the update rules of Balanced BN from TRIBE~\cite{Su2024} to update the statistics of BDN.
For a sample with pseudo-label domain $d$ and class $k$, the update rules are simplified as follows:
\begin{align}
    \bm{u}_{dk} &\leftarrow (1-\eta)\bm{u}_{dk} + \eta \overline{F_{c,\cdot,\cdot}} \\
    \bm{\sigma}^2_{dk} &\leftarrow (1-\eta)\bm{\sigma}^2_{dk} + \eta \overline{(F_{c,\cdot,\cdot} - \bm{u}_{dk})^2} - \eta^2 (\overline{F_{c,\cdot,\cdot}} - \bm{u}_{dk})^2 \\
    \bm{\mu}_d &\leftarrow \overline{\bm{u}_{d\cdot}} \\
    \bm{\sigma}^2_d &\leftarrow \overline{\bm{\sigma}^2_{d\cdot}} + \overline{(\bm{u}_{d\cdot} - \bm{\mu}_d)^2}
\end{align}
where the momentum coefficient $\eta$ is set to $5 \times 10^{-4} \times K_C$ following TRIBE~\cite{Su2024} and $K_C$ is the number of classes.

\section{Appendix: Experiments}
\label{sec:appendix_exp}

This section provides the detailed experimental setup,  additional results on all $24$ settings of the \name~benchmark, hyperparameter sensitivity analysis and visualization.

\subsection{Experimental Setup}
\label{sec:appendix_exp_setup}

We conduct experiments on three test-time adaptation datasets: CIFAR10-C~\cite{Hendrycks2019}, CIFAR100-C~\cite{Hendrycks2019}, and ImageNet-C~\cite{Hendrycks2019}. Each dataset includes 15 different corruptions at 5 levels of severity.
We evaluate all methods under the highest corruption severity level, level 5. 
Following previous works~\cite{Wang2021, Wang2022, Yuan2023, Su2024}, we adopt a standard pre-trained WideResNet-28~\cite{Zagoruyko2016}, ResNeXt-29~\cite{Xie2017}, and ResNet-50~\cite{He2016} as the backbone networks for CIFAR10-C, CIFAR100-C, and ImageNet-C, respectively.
The batch size is set to 64 for CIFAR10-C and CIFAR100-C, and 32 for ImageNet-C.
For all comparison methods, we use the original optimizers, learning rate schedules, and hyperparameter settings as described in the respective papers. 
All experiments are conducted on a single NVIDIA GeForce RTX 3090 GPU.

\begin{figure}[H]
	\centering
 	\begin{minipage}{0.45\linewidth}
		\centering
        \includegraphics[width=1.0\linewidth]{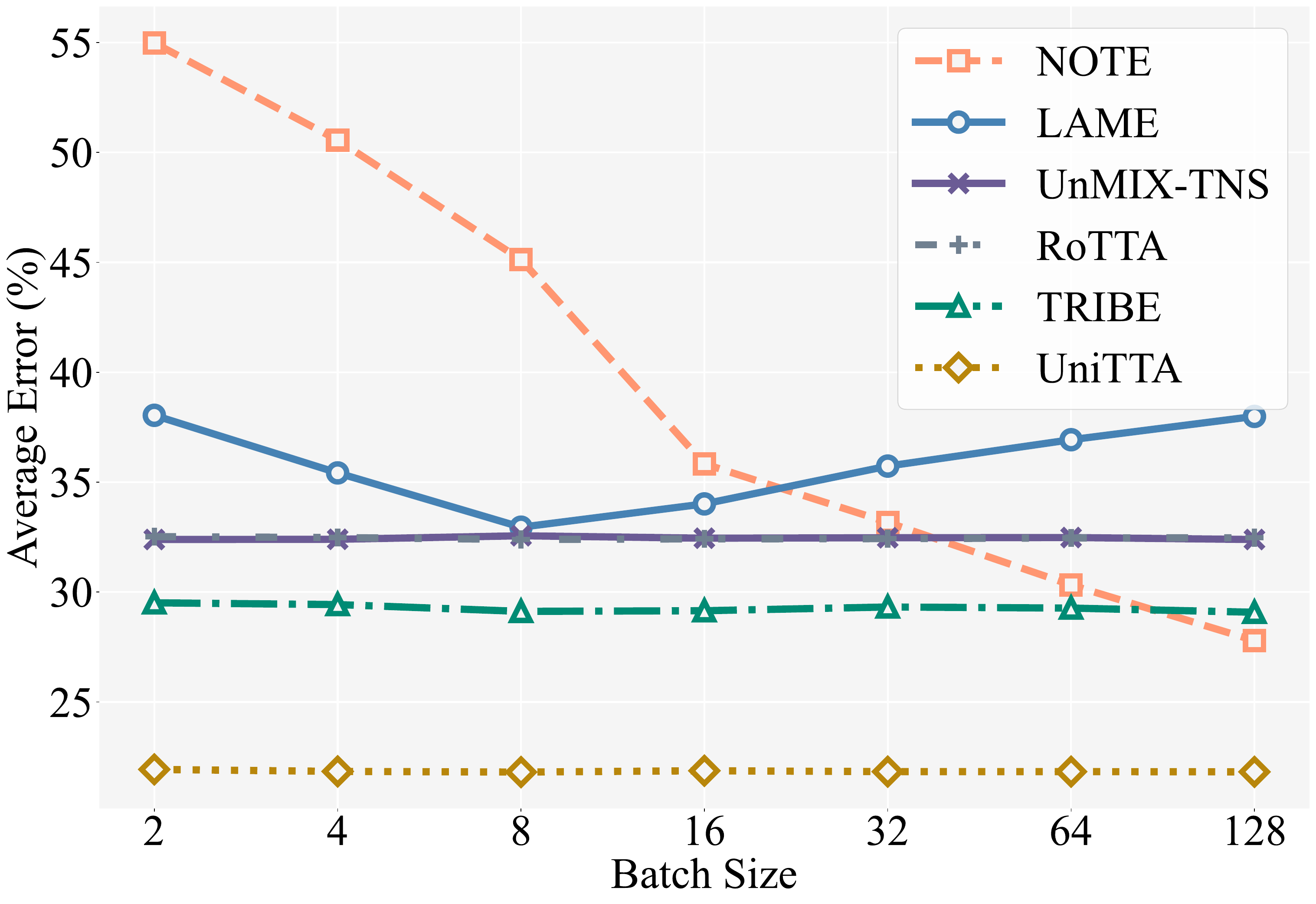}
        \caption{Sensitive analysis of batch size on CIFAR10-C. The default correlation and imbalance factors for domain and class are $(0.85, 5)$ and $(0.95, 10)$, rspectively.}

        \label{fig:bs}
	\end{minipage}
 \hfill
    \begin{minipage}{0.45\linewidth}
        \centering
        \includegraphics[width=1.0\linewidth]{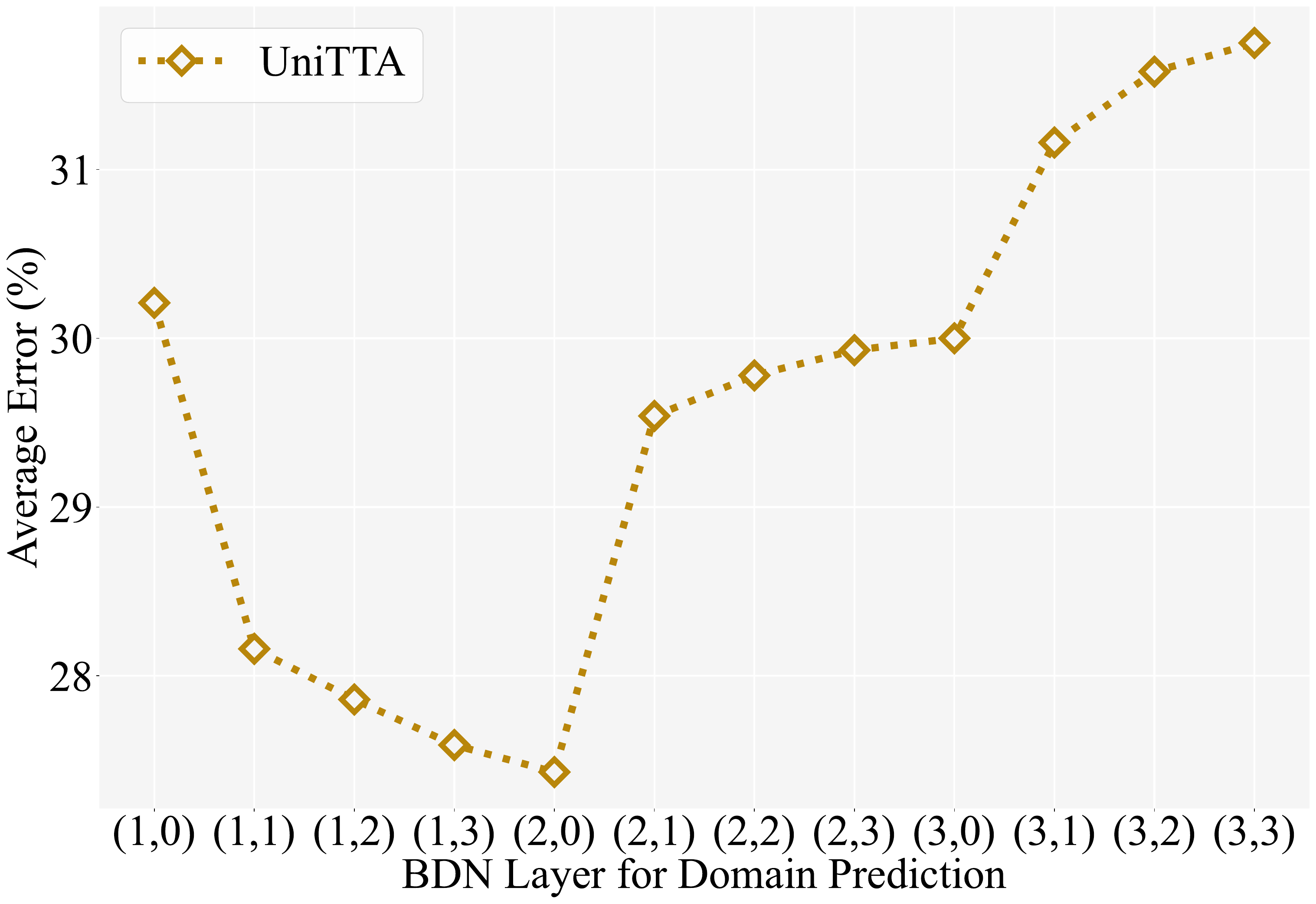}
        \caption{Sensitivity analysis of the BDN layer for domain prediction on CIFAR100-C. The horizontal axis $(m,n)$ indicates the $n$th layer of the $m$th block in the network.}
        \label{fig:bn}
	\end{minipage}
\end{figure}

For our \name~framework, mainly following the results of \cref{fig:bn}, we set the BDN layer for domain prediction to the block2.layer.0.bn1, stage2.0.bn and  layer3.0.bn1 for WideResNet-28, ResNeXt-29, and ResNet-50, respectively.
For all settings of the \name~benchmark, unless otherwise specified, the correlation factor $\alpha_1$ of non-i.i.d. settings for the domain and class is $0.85$ and $0.95$, respectively.
The imbalance factor $\beta$ for the domain and class is $5$ and $10$, respectively.
The correlation factor $\alpha_1$ is $1/K$ for the i.i.d. settings, where $K$ is the number of classes or domains.
For the balanced settings, the imbalance factor $\beta$ is $1$.
For the continual settings, the correlation factor $\alpha_1$ is $1$.

\subsection{Hyperparameter Sensitivity}
\label{sec:appendix_hyper}

\subsection{Visualization of Dynamic Domain Expansion.}
\label{sec:appendix_vis}

\begin{figure}[H]
	\centering
    \includegraphics[width=0.5\linewidth]{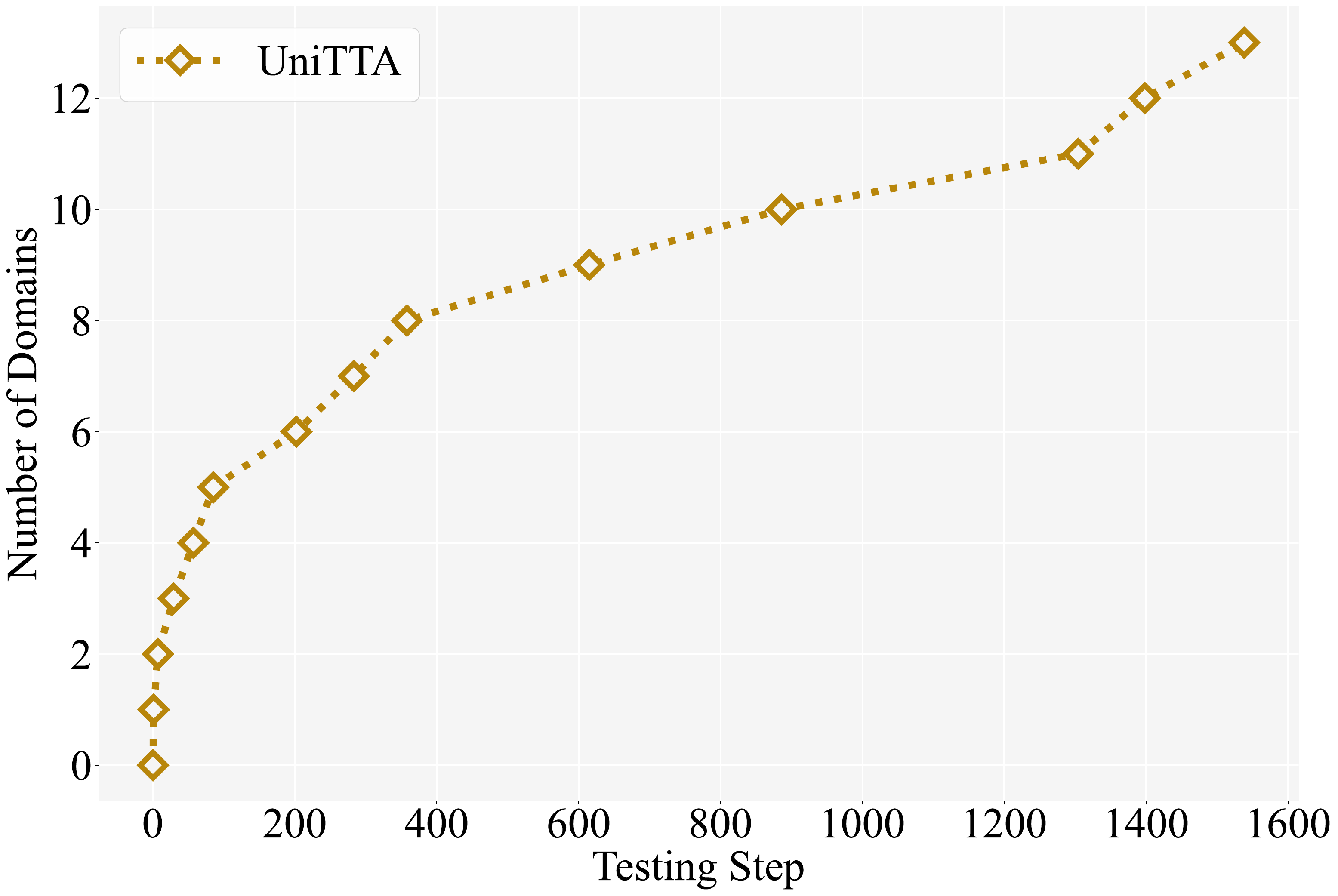}
    \caption{Visualization of dynamic domain expansion on CIFAR10-C. The BDN layer dynamically expands the domains based on the KL divergence of the domain-wise statistics.
    Only domains with more than $100$ samples are counted.}
   \label{fig:vis}
\end{figure}

\newpage

\subsection{Results on All 24 Settings}
\label{sec:appendix_more_results}

\begin{table*}[h]
    \caption{\textbf{Average error ($\%$) on CIFAR10-C within the \name~benchmark.}
    $(\{i,n,1\},\{1,u\})$ denotes correlation and imbalance settings, where $\{i,n,1\}$ represent i.i.d., non-i.i.d. and continual, respectively, and $\{1,u\}$ represent balance and imbalance, respectively.
Corresponding setting denotes the existing setting and method as shown in \cref{tab:benchmark}.}
\label{tab:all_cifar10_1}
\centering
 \resizebox{1.0\linewidth}{!}{%
     \begin{tabular}{lcccccccccccc}

     \toprule

     Class setting   & \multicolumn{6}{c}{non-i.i.d. and balanced (i,1)} & \multicolumn{6}{c}{non-i.i.d. and imbalanced (i,u)} \\

     \cmidrule(r){1-1} \cmidrule(lr){2-7} \cmidrule(l){8-13} \\

     Domain setting    &  (1,1) & (i,1) & (i,u) & (n,1) & (n,u) & (1,u) & (1,1) & (i,1) & (i,u) & (n,1) & (n,u) & (1,u)  \\

     \cmidrule(r){1-1} \cmidrule(lr){2-7} \cmidrule(l){8-13} \\
     Corresponding setting &  RoTTA  & -- & -- & -- & -- & -- & TRIBE & -- & -- & -- & -- & -- \\
     \cmidrule(r){1-1} \cmidrule(lr){2-7} \cmidrule(l){8-13} \\

    TENT~\cite{Wang2021}   & {70.29} & {83.23} & {73.16} & {78.79} & {69.18} & {60.13} & {47.40} & {59.57} & {51.48} & {62.00} & {51.90} & {44.89} \\
    TEST                    & {43.76} & {43.52} & {40.37} & {43.45} & {40.68} & {40.30} & {42.46} & {42.83} & {38.74} & {42.29} & {39.39} & {38.77} \\
    LAME~\cite{Boudiaf2022}   &  {41.40} & {40.48} & {36.98} & {40.15} & {37.49} & {37.62} & {40.91} & {40.64} & {36.58} & {40.16} & {36.93} & {36.88} \\
    ROID~\cite{Marsden2024}  &  {43.79} & {56.93} & {53.78} & {53.55} & {52.75} & {42.84} & {41.35} & {52.55} & {48.47} & {51.43} & {48.84} & {40.08} \\
    CoTTA~\cite{Wang2022}       & {53.21} & {63.93} & {62.77} & {61.67} & {61.21} & {51.96} & {41.46} & {55.18} & {50.27} & {53.20} & {50.42} & {40.42} \\
    BN~\cite{Nado2020}          & {49.42} & {57.00} & {54.82} & {56.26} & {54.91} & {48.47} & {41.52} & {50.65} & {46.52} & {50.08} & {47.27} & {39.94}\\
    Robust BN~\cite{Yuan2023}   & {23.34} & {35.61} & {31.99} & {36.04} & {32.44} & {22.01} & {26.52} & {38.52} & {34.09} & {39.63} & {35.16} & {24.78} \\
    UnMIX-TNS~\cite{Tomar2024}  & {24.68} & {32.99} & {29.03} & {32.72} & {29.15} & {25.25} & {27.60} & {35.81} & {31.88} & {36.44} & {32.48} & {27.48} \\
    Balanced BN~\cite{Su2024}  & {21.37} & {34.13} & {30.28} & {34.25} & {30.71} & {20.04} & {22.25} & {34.79} & {31.02} & {35.68} & {31.64} & {20.54} \\
    RoTTA~\cite{Yuan2023}     & {19.52} & {36.89} & {31.49} & {35.66} & {31.58} & {20.51} & {20.39} & {36.24} & {31.67} & {36.33} & {32.46} & {20.53} \\
    NOTE~\cite{Gong2022}     & {31.79} & {38.52} & {27.58} & {32.98} & {28.49} & {26.28} & {34.92} & {34.79} & {28.58} & {33.99} & {30.32} & {29.28} \\
    TRIBE~\cite{Su2024}     & {18.54} & {32.37} & {28.34} & {32.57} & {28.87} & {17.75} & \textbf{17.75} & {32.60} & {28.69} & {32.92} & {29.32} & {16.87} \\
    \midrule
    \lorename (w/o filter)      & {37.63} & {31.19} & {28.33} & {36.26} & {32.10} & {33.70} & {37.91} & {32.43} & {29.05} & {36.74} & {32.68} & {33.83} \\
    \lorename                  & {35.65} & {32.66} & {31.11} & {35.87} & {32.65} & {33.84} & {37.09} & {34.21} & {33.04} & {37.70} & {33.41} & {33.73} \\
    BDN (w/o filter)     & {24.71} & {27.46} & {24.62} & {27.74} & {24.49} & {22.71} & {25.57} & {29.64} & {26.04} & {29.62} & {26.97} & {23.38} \\
    BDN                 & {21.22} & {28.16} & {25.02} & {28.36} & {24.85} & {19.42} & {22.53} & {30.18} & {26.40} & {29.92} & {27.01} & {20.50} \\
    \cellcolor{lightgray!50}\name &\cellcolor{lightgray!50}\textbf{16.40} & \cellcolor{lightgray!50}\textbf{18.53} & \cellcolor{lightgray!50}\textbf{16.19} & \cellcolor{lightgray!50}\textbf{20.09} & \cellcolor{lightgray!50}\textbf{17.20} & \cellcolor{lightgray!50}\textbf{15.34} & \cellcolor{lightgray!50}{17.93} & \cellcolor{lightgray!50}\textbf{20.88} & \cellcolor{lightgray!50}\textbf{18.46} & \cellcolor{lightgray!50}\textbf{22.89} & \cellcolor{lightgray!50}\textbf{19.88} & \cellcolor{lightgray!50}\textbf{16.41} \\
   \bottomrule
    \end{tabular}
}
\end{table*}

\begin{table*}[h]
    \caption{\textbf{Average error ($\%$) on CIFAR10-C within the \name~benchmark.} Continuation of the previous table. "Avg." represents the average error rate across 24 settings.}
\label{tab:all_cifar10_2}
\centering
 \resizebox{1.0\linewidth}{!}{%
     \begin{tabular}{lccccccccccccl}

     \toprule

     Class setting   & \multicolumn{6}{c}{i.i.d. and balanced (i,1)} & \multicolumn{6}{c}{i.i.d. and imbalanced (i,u)} & \\

     \cmidrule(r){1-1} \cmidrule(lr){2-7} \cmidrule(lr){8-13} \\

     Domain setting    &  (1,1) & (i,1) & (i,u) & (n,1) & (n,u) & (1,u) & (1,1) & (i,1) & (i,u) & (n,1) & (n,u) & (1,u) & \\

     \cmidrule(r){1-1} \cmidrule(lr){2-7} \cmidrule(lr){8-13} \\
     Corresponding setting &  CoTTA  & ROID & -- & -- & -- & -- & -- & -- & -- & -- & -- & -- & Avg. \\
     \cmidrule(r){1-1} \cmidrule(lr){2-7} \cmidrule(lr){8-13}  \cmidrule(l){14-14} \\
     TENT~\cite{Wang2021}   & {24.03} & {59.37} & {38.58} & {47.07} & {37.81} & {20.88} & {23.36} & {48.18} & {39.40} & {36.45} & {32.34} & {22.03} & {49.23} \\
    TEST                    & {43.46} & {43.45} & {40.30} & {43.52} & {40.49} & {40.22} & {42.46} & {42.83} & {38.93} & {42.52} & {39.53} & {38.82} & {41.38} \\
    LAME~\cite{Boudiaf2022}     &  {45.07} & {44.60} & {41.35} & {44.94} & {41.62} & {41.68} & {42.92} & {42.93} & {39.02} & {42.79} & {39.58} & {39.15} & {40.49} \\
    ROID~\cite{Marsden2024}     &  {16.92} & {31.00} & {27.39} & {28.03} & {26.13} & \textbf{15.71} & {34.13} & {45.68} & {41.53} & {43.83} & {41.20} & {32.59} & {40.44} \\
    CoTTA~\cite{Wang2022}       &  \textbf{16.68} & {33.76} & {28.75} & {27.67} & {25.52} & {15.99} & {18.86} & {35.18} & {33.01} & {33.39} & {35.08} & {18.64} & {40.34} \\
    BN~\cite{Nado2020}          & {21.00} & {34.18} & {30.23} & {32.12} & {29.39} & {18.78} & {26.18} & {38.46} & {34.04} & {36.13} & {34.00} & {23.78} & {39.80} \\
    Robust BN~\cite{Yuan2023}   & {20.90} & {33.81} & {29.89} & {34.30} & {30.17} & {19.35} & {26.00} & {38.25} & {33.86} & {38.28} & {34.64} & {24.00} & {30.98} \\
    UnMIX-TNS~\cite{Tomar2024}  & {24.53} & {32.82} & {28.80} & {32.98} & {28.91} & {24.85} & {27.59} & {35.82} & {31.74} & {35.82} & {32.54} & {27.50} & {30.39} \\
    Balanced BN~\cite{Su2024}  & {21.18} & {33.85} & {30.03} & {34.31} & {30.22} & {19.60} & {22.25} & {34.97} & {30.92} & {34.82} & {31.52} & {20.31} & {28.78} \\
    RoTTA~\cite{Yuan2023}       & {17.84} & {33.45} & {29.50} & {33.58} & {29.73} & {18.78} & {18.88} & {35.62} & {31.21} & {35.19} & {31.79} & {19.32} & {28.67} \\
    NOTE~\cite{Gong2022}        & {22.55} & \textbf{24.48} & \textbf{22.33} & \textbf{24.06} & \textbf{22.35} & {21.85} & {26.39} & {30.62} & \textbf{25.87} & \textbf{29.37} & \textbf{26.48} & {24.79} & {28.28} \\
    TRIBE~\cite{Su2024}         & {18.20} & {31.90} & {27.97} & {32.29} & {28.04} & {17.29} & \textbf{17.77} & {32.71} & {28.19} & {31.96} & {28.82} & \textbf{16.47} & {26.18} \\
    \midrule
    \lorename (w/o filter)      & {65.98} & {62.97} & {61.96} & {65.45} & {63.52} & {64.31} & {63.74} & {60.90} & {59.57} & {63.22} & {60.88} & {61.65} & {48.17} \\
    \lorename                   & {47.75} & {46.71} & {43.82} & {47.59} & {44.64} & {44.58} & {46.18} & {45.41} & {41.99} & {46.08} & {43.15} & {42.65} & {40.06} \\
    BDN (w/o filter)       & {24.71} & {27.46} & {24.62} & {27.74} & {24.49} & {22.71} & {25.57} & \textbf{29.64} & {26.04} & {29.62} & {26.97} & {23.38} & {26.35} \\
    BDN                    & {21.22} & {28.16} & {25.02} & {28.36} & {24.85} & {19.42} & {22.53} & {30.18} & {26.40} & {29.92} & {27.01} & {20.50} & {25.63} \\
    \cellcolor{lightgray!50}\name &\cellcolor{lightgray!50}{28.38} & \cellcolor{lightgray!50}{31.34} & \cellcolor{lightgray!50}{28.44} & \cellcolor{lightgray!50}{31.81} & \cellcolor{lightgray!50}{28.58} & \cellcolor{lightgray!50}{26.25} & \cellcolor{lightgray!50}{28.89} & \cellcolor{lightgray!50}{32.39} & \cellcolor{lightgray!50}{28.99} & \cellcolor{lightgray!50}{32.77} & \cellcolor{lightgray!50}{30.31} & \cellcolor{lightgray!50}{26.54} & \cellcolor{lightgray!50}\textbf{23.95 \color{teal}{(-2.23)}} \\

   \bottomrule
    \end{tabular}
}
\end{table*}

\begin{table*}[!t]
    \caption{\textbf{Average error ($\%$) on CIFAR100-C within the \name~benchmark.}
    $(\{i,n,1\},\{1,u\})$ denotes correlation and imbalance settings, where $\{i,n,1\}$ represent i.i.d., non-i.i.d. and continual, respectively, and $\{1,u\}$ represent balance and imbalance, respectively.
Corresponding setting denotes the existing setting and method as shown in \cref{tab:benchmark}.}
    \label{tab:all_cifar100_1}
    \centering
    \resizebox{1.0\linewidth}{!}{%
        \begin{tabular}{lcccccccccccc}

        \toprule
        Class setting   & \multicolumn{6}{c}{non-i.i.d. and balanced (i,1)} & \multicolumn{6}{c}{non-i.i.d. and imbalanced (i,u)} \\

        \cmidrule(r){1-1} \cmidrule(lr){2-7} \cmidrule(l){8-13} \\

        Domain setting    &  (1,1) & (i,1) & (i,u) & (n,1) & (n,u) & (1,u) & (1,1) & (i,1) & (i,u) & (n,1) & (n,u) & (1,u)  \\

        \cmidrule(r){1-1} \cmidrule(lr){2-7} \cmidrule(l){8-13} \\
        Corresponding setting &  RoTTA  & -- & -- & -- & -- & -- & TRIBE & -- & -- & -- & -- & -- \\
        \cmidrule(r){1-1} \cmidrule(lr){2-7} \cmidrule(l){8-13} \\

        TENT~\cite{Wang2021}   & {96.53} & {97.08} & {94.26} & {95.08} & {89.75} & {94.91} & {93.79} & {93.74} & {86.80} & {88.26} & {83.97} & {90.80}  \\
        NOTE~\cite{Gong2022}      & {79.69} & {67.67} & {57.69} & {59.75} & {54.37} & {65.43} & {71.52} & {58.86} & {55.52} & {57.55} & {54.25} & {61.70} \\
        BN~\cite{Nado2020}     & {76.55} & {79.33} & {78.55} & {79.15} & {79.42} & {76.22} & {64.42} & {69.33} & {69.68} & {69.11} & {68.49} & {63.69}  \\
        CoTTA~\cite{Wang2022}  & {78.26} & {79.95} & {78.91} & {78.89} & {79.38} & {76.77} & {65.68} & {68.56} & {69.58} & {68.07} & {68.46} & {63.95} \\
        ROID~\cite{Marsden2024} & {71.09} & {77.57} & {76.80} & {76.14} & {76.47} & {70.56} & {55.27} & {63.21} & {63.88} & {62.70} & {63.38} & {54.83} \\
        RoTTA~\cite{Yuan2023}    & {38.95} & {53.80} & {52.30} & {53.25} & {52.55} & {40.44} & {37.79} & {54.99} & {53.89} & {55.36} & {53.63} & {40.34}  \\
        TEST                  & {46.64} & {46.66} & {45.11} & {47.33} & {44.89} & {45.11} & {47.07} & {46.86} & {45.83} & {47.87} & {46.05} & {45.04}  \\
        Robust BN~\cite{Yuan2023} & {40.90} & {50.09} & {48.75} & {51.17} & {49.13} & {40.36} & {39.33} & {48.50} & {48.14} & {49.90} & {48.48} & {38.64}  \\
        UnMIX-TNS~\cite{Tomar2024} & {39.12} & {46.88} & {45.66} & {46.92} & {45.36} & {40.19} & {40.19} & {47.44} & {46.41} & {47.55} & {46.20} & {41.00}  \\
        Balanced BN~\cite{Su2024}  & {36.36} & {46.47} & {45.66} & {47.01} & {45.16} & {36.47} & {36.77} & {46.67} & {46.39} & {47.30} & {46.40} & {36.47}  \\
        TRIBE~\cite{Su2024}      & {34.69} & {47.95} & {43.75} & {47.89} & {44.37} & {35.02} & {32.74} & {46.67} & {43.19} & {46.35} & {44.37} & {32.83}  \\
        \midrule
        \lorename (w/o filter)      & {32.88} & \textbf{28.52} & \textbf{26.98} & {32.69} & \textbf{28.98} & {30.96} & {36.04} & {31.89} & \textbf{30.58} & {36.71} & {33.45} & {34.05}  \\
        \lorename                  & {35.65} & {32.66} & {31.11} & {35.87} & {32.65} & {33.84} & {37.09} & {34.21} & {33.04} & {37.70} & {34.97} & {34.94} \\
        BDN (w/o filter)     & {38.85} & {44.35} & {43.52} & {44.99} & {43.62} & {39.62} & {38.58} & {44.37} & {44.38} & {44.74} & {44.53} & {38.48}  \\
        BDN                 & {36.19} & {43.46} & {42.70} & {44.10} & {42.86} & {36.37} & {36.03} & {43.50} & {43.41} & {43.63} & {43.34} & {35.91} \\
        \cellcolor{lightgray!50}\name &\cellcolor{lightgray!50}\textbf{24.49} & \cellcolor{lightgray!50}{28.99} & \cellcolor{lightgray!50}{28.57} & \cellcolor{lightgray!50}\textbf{31.85} & \cellcolor{lightgray!50}{29.53} & \cellcolor{lightgray!50}\textbf{25.11} & \cellcolor{lightgray!50}\textbf{25.81} & \cellcolor{lightgray!50}\textbf{30.96} & \cellcolor{lightgray!50}{30.95} & \cellcolor{lightgray!50}\textbf{32.87} & \cellcolor{lightgray!50}\textbf{32.16} & \cellcolor{lightgray!50}\textbf{26.26} \\

        \bottomrule
        \end{tabular}
    }

\end{table*}

\begin{table*}[!t]
    \caption{\textbf{Average error ($\%$) on CIFAR100-C within the \name~benchmark.} Continuation of the previous table. "Avg." represents the average error rate across 24 settings.}
\label{tab:all_cifar100_2}
\centering
 \resizebox{1.0\linewidth}{!}{%
     \begin{tabular}{lccccccccccccl}

     \toprule

     Class setting   & \multicolumn{6}{c}{i.i.d. and balanced (i,1)} & \multicolumn{6}{c}{i.i.d. and imbalanced (i,u)} \\

     \cmidrule(r){1-1} \cmidrule(lr){2-7} \cmidrule(l){8-13} \\

     Domain setting    &  (1,1) & (i,1) & (i,u) & (n,1) & (n,u) & (1,u) & (1,1) & (i,1) & (i,u) & (n,1) & (n,u) & (1,u) \\

     \cmidrule(r){1-1} \cmidrule(lr){2-7} \cmidrule(l){8-13} \\
     Corresponding setting &  CoTTA  & ROID & -- & -- & -- & -- & -- & -- & -- & -- & -- & -- & Avg. \\
     \cmidrule(r){1-1} \cmidrule(lr){2-7} \cmidrule(l){8-13} \\
     TENT~\cite{Wang2021}   & {81.06} & {91.05} & {83.37} & {88.59} & {79.70} & {63.18} & {76.04} & {73.53} & {58.98} & {53.84} & {50.11} & {60.56} & {81.87} \\
    NOTE~\cite{Gong2022}      & {65.96} & {63.07} & {56.56} & {63.47} & {56.10} & {57.57} & {67.54} & {56.62} & {54.39} & {55.59} & {52.80} & {57.30} & {60.46} \\
    BN~\cite{Nado2020}     & {36.20} & {46.48} & {44.93} & {45.04} & {44.27} & {35.34} & {37.36} & {47.70} & {46.64} & {46.45} & {44.78} & {36.53} & {57.74} \\
    CoTTA~\cite{Wang2022}  & {32.74} & {43.02} & {42.47} & {41.22} & {41.91} & {32.61} & {33.47} & {44.37} & {45.01} & {43.60} & {43.65} & {33.56} & {56.42} \\
    ROID~\cite{Marsden2024}  & \textbf{29.91} & \textbf{36.84} & \textbf{36.65} & \textbf{36.81} & \textbf{36.71} & \textbf{29.90} & {31.89} & \textbf{38.71} & \textbf{39.31} & \textbf{38.84} & \textbf{38.42} & \textbf{31.70} & {51.57} \\
    RoTTA~\cite{Yuan2023}      & {33.46} & {46.54} & {46.63} & {47.28} & {46.46} & {35.41} & {34.00} & {51.43} & {51.30} & {53.71} & {50.63} & {36.07} & {46.68} \\
    TEST                    & {46.35} & {46.43} & {44.55} & {46.72} & {44.53} & {44.53} & {46.94} & {46.80} & {45.84} & {47.43} & {44.48} & {44.88} & {46.00} \\
    Robust BN~\cite{Yuan2023} & {35.56} & {45.99} & {44.45} & {46.64} & {44.56} & {35.18} & {36.73} & {46.97} & {46.32} & {47.91} & {44.99} & {36.29} & {44.37} \\
    UnMIX-TNS~\cite{Tomar2024} & {38.94} & {46.32} & {44.66} & {46.61} & {44.75} & {39.78} & {39.96} & {47.16} & {46.23} & {47.58} & {44.94} & {40.78} & {44.19} \\
    Balanced BN~\cite{Su2024}  & {35.84} & {45.94} & {44.38} & {46.38} & {44.43} & {35.62} & {36.32} & {46.50} & {45.82} & {46.98} & {44.71} & {35.99} & {42.75} \\
    TRIBE~\cite{Su2024}      & {33.10} & {45.73} & {42.99} & {46.84} & {43.38} & {32.99} & \textbf{31.71} & {45.28} & {43.01} & {46.60} & {41.65} & {31.82} & {41.04} \\
    LAME~\cite{Boudiaf2022}  & {48.21} & {47.47} & {45.59} & {47.90} & {45.58} & {46.34} & {48.23} & {47.34} & {46.35} & {48.00} & {45.06} & {46.00} & {40.41} \\
    \midrule
    \lorename (w/o filter)   & {70.82} & {69.50} & {68.33} & {70.65} & {69.22} & {69.52} & {70.60} & {69.43} & {68.70} & {70.66} & {68.41} & {69.45} & {50.79} \\
    \lorename                & {51.64} & {51.50} & {49.83} & {52.04} & {49.98} & {50.02} & {52.18} & {51.71} & {50.73} & {52.65} & {49.99} & {50.29} & {42.76} \\
    BDN (w/o filter)    & {37.82} & {43.70} & {42.17} & {43.56} & {41.95} & {37.42} & {37.77} & {44.20} & {43.44} & {45.02} & {42.33} & {37.86} & {41.97} \\
    BDN                 & {34.65} & {41.92} & {40.55} & {42.29} & {40.41} & {34.15} & {35.06} & {42.43} & {42.10} & {43.54} & {41.34} & {34.58} & {40.19} \\
    \cellcolor{lightgray!50}\name &\cellcolor{lightgray!50}{44.17} &  \cellcolor{lightgray!50}{48.86} &  \cellcolor{lightgray!50}{47.72} &  \cellcolor{lightgray!50}{49.36} &  \cellcolor{lightgray!50}{47.78} &  \cellcolor{lightgray!50}{44.08} &  \cellcolor{lightgray!50}{44.14} &  \cellcolor{lightgray!50}{49.18} &  \cellcolor{lightgray!50}{49.05} &  \cellcolor{lightgray!50}{50.33} &  \cellcolor{lightgray!50}{47.57} &  \cellcolor{lightgray!50}{43.72} &  \cellcolor{lightgray!50}\textbf{38.06 \color{teal}{(-2.35)}} \\

   \bottomrule
    \end{tabular}
}
\end{table*}

\begin{table*}[!t]
    \caption{\textbf{Average error ($\%$) on ImageNet-C within the \name~benchmark.}
    $(\{i,n,1\},\{1,u\})$ denotes correlation and imbalance settings, where $\{i,n,1\}$ represent i.i.d., non-i.i.d. and continual, respectively, and $\{1,u\}$ represent balance and imbalance, respectively.
    Corresponding setting denotes the existing setting and method as shown in \cref{tab:benchmark}.}
\label{tab:all_imagenet_1}
\centering
 \resizebox{1.0\linewidth}{!}{%
     \begin{tabular}{lcccccccccccc}

     \toprule
     Class setting   & \multicolumn{6}{c}{non-i.i.d. and balanced (i,1)} & \multicolumn{6}{c}{non-i.i.d. and imbalanced (i,u)} \\

     \cmidrule(r){1-1} \cmidrule(lr){2-7} \cmidrule(l){8-13} \\

     Domain setting    &  (1,1) & (i,1) & (i,u) & (n,1) & (n,u) & (1,u) & (1,1) & (i,1) & (i,u) & (n,1) & (n,u) & (1,u)  \\

     \cmidrule(r){1-1} \cmidrule(lr){2-7} \cmidrule(l){8-13} \\
     Corresponding setting &  RoTTA  & -- & -- & -- & -- & -- & TRIBE & -- & -- & -- & -- & -- \\
     \cmidrule(r){1-1} \cmidrule(lr){2-7} \cmidrule(l){8-13} \\
     NOTE~\cite{Gong2022}  &{93.67} & {95.27} & {96.82} & {95.00} & {95.81} & {88.75} & {92.49} & {95.93} & {95.41} & {88.93} & {95.05} & {85.85} \\
    TENT~\cite{Wang2021}   & {98.72} & {99.31} & {99.53} & {99.12} & {99.32} & {97.69} & {97.50} & {99.22} & {99.13} & {97.03} & {98.86} & {94.71} \\
    TRIBE~\cite{Su2024}      & {89.78} & {92.62} & {96.54} & {95.19} & {95.99} & {78.04} & {88.72} & {92.85} & {93.71} & {89.37} & {94.05} & {69.34} \\
    ROID~\cite{Marsden2024} & {98.51} & {99.71} & {99.84} & {99.52} & {99.61} & {97.35} & {91.76} & {99.77} & {99.57} & {98.15} & {99.37} & {91.75} \\
    BN~\cite{Nado2020}     & {93.79} & {95.08} & {95.15} & {95.10} & {95.01} & {93.76} & {88.40} & {92.24} & {92.25} & {91.31} & {91.84} & {88.34} \\
    CoTTA~\cite{Wang2022}  & {95.13} & {96.80} & {97.33} & {96.22} & {96.33} & {94.59} & {89.70} & {95.20} & {94.50} & {92.11} & {93.71} & {89.04} \\
    Robust BN~\cite{Yuan2023} & {80.76} & {89.58} & {89.58} & {91.74} & {90.40} & {81.08} & {74.69} & {87.16} & {87.31} & {89.24} & {88.46} & {75.82} \\
    UnMIX-TNS~\cite{Tomar2024} & {79.74} & {84.42} & {82.67} & {84.57} & {82.91} & {82.08} & {78.67} & {83.28} & {82.34} & {85.04} & {82.38} & {81.52} \\
    TEST                    & {81.92} & {82.10} & {81.66} & {81.96} & {81.74} & {82.07} & {81.60} & {81.21} & {81.42} & {81.20} & {81.52} & {81.95} \\
    Balanced BN~\cite{Su2024}  & {76.63} & {87.03} & {86.54} & {88.87} & {87.23} & {77.24} & \textbf{71.19} & {84.38} & {84.41} & {86.22} & {85.35} & {72.27} \\
    LAME~\cite{Boudiaf2022}  & {74.48} & {72.21} & {71.77} & {73.52} & {73.13} & {74.69} & {75.70} & {73.44} & {73.54} & {74.38} & {74.39} & {76.25}\\
    RoTTA~\cite{Yuan2023}      & {71.72} & {80.54} & {79.65} & {80.30} & {79.63} & {73.59} & {68.74} & {78.26} & {77.94} & {79.78} & {78.36} & {72.47} \\
    \midrule
    \lorename (w/o filter)   & {75.37} & {70.61} & {69.75} & {74.42} & {73.22} & {75.30} & {76.32} & {71.07} & {70.57} & {75.06} & {74.56} & {76.67} \\
    \lorename               & {76.62} & {73.86} & {73.51} & {76.17} & {75.41} & {76.97} & {76.82} & {73.28} & {73.59} & {75.82} & {75.77} & {77.29} \\
    BDN (w/o filter)  & {77.80} & {76.37} & {76.03} & {76.88} & {76.38} & {79.21} & {76.69} & {75.13} & {75.74} & {75.97} & {75.99} & {78.48} \\
    BDN              & {76.69} & {79.48} & {79.32} & {79.82} & {79.28} & {77.68} & {72.87} & {77.83} & {78.21} & {77.89} & {78.12} & {74.16} \\
    \cellcolor{lightgray!50}\name &\cellcolor{lightgray!50}\textbf{70.25} &  \cellcolor{lightgray!50}\textbf{66.83} &  \cellcolor{lightgray!50}\textbf{66.42} &  \cellcolor{lightgray!50}\textbf{68.29} &  \cellcolor{lightgray!50}\textbf{68.05} &  \cellcolor{lightgray!50}\textbf{72.39} &  \cellcolor{lightgray!50}{72.02} &  \cellcolor{lightgray!50}\textbf{65.68} &  \cellcolor{lightgray!50}\textbf{66.87} &  \cellcolor{lightgray!50}\textbf{68.48} &  \cellcolor{lightgray!50}\textbf{67.58} &  \cellcolor{lightgray!50}\textbf{71.70} \\
   \bottomrule
    \end{tabular}
}
\end{table*}

\begin{table*}[!t]
    \caption{\textbf{Average error ($\%$) on ImageNet-C within the \name~benchmark.}
    Continuation of the previous table. "Avg." represents the average error rate across 24 settings.}
\label{tab:all_imagenet_2}
\centering
 \resizebox{1.0\linewidth}{!}{%
     \begin{tabular}{lccccccccccccl}

     \toprule
     Class setting   & \multicolumn{6}{c}{i.i.d. and balanced (i,1)} & \multicolumn{6}{c}{i.i.d. and imbalanced (i,u)} \\

     \cmidrule(r){1-1} \cmidrule(lr){2-7} \cmidrule(lr){8-13} \\

     Domain setting    &  (1,1) & (i,1) & (i,u) & (n,1) & (n,u) & (1,u) & (1,1) & (i,1) & (i,u) & (n,1) & (n,u) & (1,u) \\

     \cmidrule(r){1-1} \cmidrule(lr){2-7} \cmidrule(lr){8-13} \\
     Corresponding setting &  CoTTA  & ROID & -- & -- & -- & -- & -- & -- & -- & -- & -- & -- & Avg. \\
     \cmidrule(r){1-1} \cmidrule(lr){2-7} \cmidrule(lr){8-13} \cmidrule(l){14-14} \\
     NOTE~\cite{Gong2022}  & {91.62} & {88.18} & {83.53} & {86.89} & {87.55} & {85.78} & {88.90} & {94.75} & {94.02} & {91.06} & {94.64} & {83.18} & {91.21} \\
     TENT~\cite{Wang2021}   & {70.58} & {91.88} & {82.37} & {80.13} & {85.00} & {64.92} & {68.21} & {97.29} & {96.06} & {87.28} & {94.29} & \textbf{62.02} & {90.01} \\
    TRIBE~\cite{Su2024}      & {75.85} & {84.78} & {83.59} & {84.61} & {83.61} & {63.98} & {79.88} & {85.48} & {87.38} & {87.02} & {84.86} & {62.18} & {84.98} \\
ROID~\cite{Marsden2024} & {60.67} & {79.18} & {83.83} & {77.54} & {78.46} & \textbf{62.25} & \textbf{57.70} & {76.61} & {76.82} & {73.52} & {76.20} & {58.95} & {84.86} \\
    BN~\cite{Nado2020}     & {69.33} & {82.87} & {83.22} & {79.40} & {80.45} & {69.44} & {67.57} & {81.79} & {81.58} & {77.73} & {79.33} & {68.19} & {84.71} \\
    CoTTA~\cite{Wang2022}  & \textbf{66.87} & {80.67} & {82.07} & \textbf{76.28} & {76.81} & {66.13} & {64.31} & {79.42} & {78.28} & \textbf{72.69} & {75.74} & {63.42} & {83.89} \\
    Robust BN~\cite{Yuan2023} & {69.81} & {84.90} & {85.16} & {87.35} & {85.64} & {70.55} & {68.37} & {84.37} & {84.17} & {86.22} & {85.65} & {69.36} & {82.81} \\ 
    UnMIX-TNS~\cite{Tomar2024} & {79.64} & {85.55} & {88.41} & {86.68} & {84.48} & {81.81} & {78.15} & {82.91} & {81.91} & {83.01} & {82.12} & {81.08} & {82.72} \\ 
    TEST                    & {81.99} & {82.05} & {83.46} & {82.78} & {82.15} & {82.14} & {80.93} & {81.00} & {81.15} & {80.69} & {81.40} & {81.17} & {81.72} \\ 
    Balanced BN~\cite{Su2024}  & {69.31} & {83.35} & {84.71} & {85.40} & {83.60} & {69.89} & {67.28} & {82.07} & {82.17} & {83.51} & {82.90} & {68.46} & {80.42} \\ 
    LAME~\cite{Boudiaf2022}  & {82.55} & {82.26} & {83.83} & {83.05} & {82.41} & {82.68} & {81.42} & {81.13} & {81.30} & {80.98} & {81.65} & {81.75} & {78.02} \\ 
    RoTTA~\cite{Yuan2023}      & {67.77} & {79.91} & {81.22} & {81.11} & {79.81} & {71.98} & {66.16} & {75.81} & {75.71} & {77.65} & {76.36} & {71.26} & {76.07} \\ 
    \midrule
    \lorename (w/o filter)   & {91.83} & {89.69} & {90.93} & {91.95} & {91.15} & {92.01} & {91.32} & {88.99} & {89.19} & {90.75} & {90.93} & {91.64} & {82.22} \\ 
    \lorename               & {82.99} & {82.77} & {84.07} & {83.52} & {83.26} & {83.24} & {82.03} & {81.81} & {81.95} & {81.79} & {82.46} & {82.19} & {79.05} \\ 
    BDN (w/o filter)   & {77.09} & {76.64} & \textbf{80.28} & {77.31} & {76.84} & {78.09} & {75.79} & \textbf{74.93} & \textbf{75.20} & {75.72} & {75.88} & {77.33} & {76.74} \\ 
    BDN              & {68.62} & \textbf{77.64} & {80.83} & {76.68} & \textbf{76.54} & {68.96} & {66.64} & {75.76} & {76.02} & {74.66} & \textbf{75.56} & {67.43} & {75.69} \\
    \cellcolor{lightgray!50}\name &\cellcolor{lightgray!50}{78.07} &  \cellcolor{lightgray!50}{78.00} &  \cellcolor{lightgray!50}{80.89} &  \cellcolor{lightgray!50}{78.32} &  \cellcolor{lightgray!50}{77.94} &  \cellcolor{lightgray!50}{79.28} &  \cellcolor{lightgray!50}{76.76} &  \cellcolor{lightgray!50}{75.96} &  \cellcolor{lightgray!50}{76.71} &  \cellcolor{lightgray!50}{76.45} &  \cellcolor{lightgray!50}{76.91} &  \cellcolor{lightgray!50}{78.30} &  \cellcolor{lightgray!50}\textbf{73.26 \color{teal}{(-2.81)}} \\

   \bottomrule
    \end{tabular}
}
\end{table*}